# Can we integrate spatial verification methods into neural-network loss functions for atmospheric science?

**This article has been accepted to the AMS journal *Artificial Intelligence for the Earth Systems*.**

Ryan Lagerquist[a,b] and Imme Ebert-Uphoff[a,c]

[a] *Cooperative Institute for Research in the Atmosphere (CIRA)*

[b] *National Oceanic and Atmospheric Administration (NOAA) Earth System Research Laboratory (ESRL) / Global Systems Laboratory (GSL), Boulder, Colorado*

[c] *Department of Electrical and Computer Engineering, Colorado State University, Fort Collins, Colorado*

*Corresponding author*: Ryan Lagerquist, ralager@colostate.edu

ABSTRACT: In the last decade, much work in atmospheric science has focused on spatial verification (SV) methods for gridded prediction, which overcome serious disadvantages of pixelwise verification. However, neural networks (NN) in atmospheric science are almost always trained to optimize pixelwise loss functions, even when ultimately assessed with SV methods. This establishes a disconnect between model verification during vs. after training. To address this issue, we develop spatially enhanced loss functions (SELF) and demonstrate their use for a real-world problem: predicting the occurrence of thunderstorms (henceforth, "convection") with NNs. In each SELF we use either a neighbourhood filter, which highlights convection at scales larger than a threshold, or a spectral filter (employing Fourier or wavelet decomposition), which is more flexible and highlights convection at scales between two thresholds. We use these filters to spatially enhance common verification scores, such as the Brier score. We train each NN with a different SELF and compare their performance at many scales of convection, from discrete storm cells to tropical cyclones. Among our many findings are that (a) for a low (high) risk threshold, the ideal SELF focuses on small (large) scales; (b) models trained with a pixelwise loss function perform surprisingly well; (c) however, models trained with a spectral filter produce much better-calibrated probabilities than a pixelwise model. We provide a general guide to using SELFs, including technical challenges and the final Python code, as well as demonstrating their use for the convection problem. To our knowledge this is the most in-depth guide to SELFs in the geosciences.

SIGNIFICANCE STATEMENT: Gridded predictions, where a quantity is predicted at every pixel in space, should be verified with spatially aware methods rather than pixel-by-pixel. Neural networks (NN), which are often used for gridded prediction, are trained to minimize an error value called the loss function. NN loss functions in atmospheric science are almost always pixelwise, which causes the predictions to miss rare events and contain unrealistic spatial patterns. We use spatial filters to enhance NN loss functions, and we test our novel spatially enhanced loss functions (SELF) on thunderstorm prediction. We find that different SELFs work better for different scales (*i.e.*, different-sized thunderstorm complexes) and that spectral filters, one of the two filter types, produce unexpectedly well calibrated thunderstorm probabilities.

# 1. Introduction

Many problems in atmospheric science involve predicting on a spatial grid. Traditionally, gridded predictions have been verified pixelwise, where the prediction (observation) at pixel $P$ is compared only to the observation (prediction) at $P$. Pixelwise verification has three main disadvantages. The first is the double-penalty problem (Gilleland et al. 2009), which is most prevalent when predicting a spatially rare event – *i.e.*, one that occurs at only a small fraction of pixels, like heavy rainfall. If the predictions have a small offset (enough that predicted and observed event areas do not line up), even if the predictions are otherwise perfect, verification scores are very poor. This is because pixels with the predicted event are counted as false positives, while pixels with the observed event are counted as false negatives. The combination of false positives and false negatives is the "double penalty," and this encourages the model not to predict spatially rare events at all. The second disadvantage is that pixelwise verification scores are not well correlated with the perceptual (*i.e.*, subjective) quality of the prediction. For example, Figure 2 of Wang and Bovik (2009) shows that pixelwise mean squared error (MSE) is often much worse for images with better perceptual quality. Third, pixelwise verification encourages models not to make predictions with sharp gradients – even if the observations contain sharp gradients – because the sharp gradients will likely not be at the exact right location, leading to high pixelwise error. By avoiding sharp gradients, models produce predictions that are too smooth or "blurry" (*e.g.*, the "CNN SR" in Figures 3 and 7 of Stengel et al. 2020).

## a. Spatial verification

To address these problems, much recent work has focused on spatial verification (SV). One of the most popular SV methods in atmospheric science is the fractions skill score (FSS; Roberts and Lean 2008), which compares gridded predictions and observations of a binary event, such as heavy rainfall, where the only possible values are 1 (yes) and 0 (no). The FSS includes a neighbourhood distance, so when focusing on pixel $P$, it actually compares the predicted and observed event fractions in the surrounding patch. This patch may be $3 \times 3$, $5 \times 5$, $7 \times 7$, etc. For a large enough patch, the FSS avoids the double-penalty problem. A similar SV method is popular in the computer-vision literature, called the structural similarity (SSIM) index (Wang et al. 2004; Wang and Bovik 2009). The SSIM index can be used to verify a regression model (one that predicts continuous values, such as rainfall amount in mm) or a classification model. At each pixel $P$, the SSIM index is the product of three similarities computed over the surrounding patch: the similarity of raw values ("luminances" in Wang and Bovik 2009), contrasts (based on the variance in each image), and structures (based on the covariance between images). The FSS has been widely adopted in atmospheric science (Weusthoff et al. 2010; Sobash et al. 2011; Mittermaier et al. 2013; Bachmann et al. 2018; Ahmed et al. 2019; Loken et al. 2019; Qian and Wang 2021), as has the SSIM index in computer science (Johnson et al. 2016; Zhao et al. 2016; Hammernik et al. 2017; Ledig et al. 2017; Snell et al. 2017; Wang et al. 2019).

SV methods in atmospheric science go far beyond the FSS. Gilleland et al. (2009) provided an overview, splitting SV methods into four categories: neighbourhood, scale-separation, feature-based, and field-deformation methods. Neighbourhood (or "fuzzy") methods, like the FSS and SSIM index, use a patch centered at pixel $P$ to compare the predictions and observations around $P$. Scale-separation methods apply a spectral filter (*i.e.*, a low-pass, high-pass, or band-pass filter implemented with either Fourier or wavelet decomposition) to both the predictions and observations, thus isolating patterns over the desired range of physical scales, followed by pixelwise verification. Feature-based methods match high-level features (*e.g.*, cold fronts) between the predictions and observations, while field-deformation (or "morphing") methods determine how much manipulation is needed to make the predictions and observations match.

*b. From spatial verification to spatially enhanced loss functions*

There is a growing recognition that physically meaningful loss functions are needed in the geosciences (Karpatne et al. 2017; Willard et al. 2020; Beucler et al. 2021; Lagerquist et al. 2021b). Neural networks (NN) have become a common tool in atmospheric science and the geosciences at large. For gridded prediction, a popular NN architecture is the U-net (Chen et al. 2020; Kumler-Bonfanti et al. 2020; Sadeghi et al. 2020; Sha et al. 2020a,b; Han et al. 2021; Lagerquist et al. 2021a,b), invented by Ronneberger et al. (2015). Lagerquist et al. (2021a) contains an explanation of U-nets for atmospheric scientists. NNs, including U-nets, are trained to minimize a loss function, which measures the error between the predictions and observations. Despite much recent work in SV, NNs in the geosciences are almost always trained with pixelwise loss functions. This establishes a disconnect between the model verification during and after training – *i.e.*, the model is trained to optimize pixelwise performance but is verified on spatial performance.

To our knowledge, spatially enhanced loss functions (SELF) appear in only a few geoscience applications. First, Zhang et al. (2008) used Baddeley's $\Delta$ metric, a field-deformation method, for a climate application: to predict the region of the Americas exceeding a given temperature change from the 1980s to 1990s. Second, Gilleland (2021) developed new field-deformation metrics and suggested their use as loss functions in models that predict a spatial-exceedance region, like that of Zhang et al. (2008). One advantage of these new metrics is that they handle the case where all values in the predicted or observed field are the same – *e.g.*, 0 for no convection. Third, Heim and Avery (2019) used the image Euclidean distance (IMED; Wang et al. 2005) to detect switches in the Kuroshio current, a major ocean current east of Japan, between an elongated and contracted state. They framed the problem as anomaly detection in time series and used a convolutional neural network to solve the problem. The IMED is similar to the pixelwise MSE but accounts for spatial autocorrelation between nearby pixels. Fourth, Stengel et al. (2020) used a composite loss function to upsample fields of wind velocity and solar irradiance. Their composite loss function is pixelwise MSE plus "adversarial loss," which is provided by the discriminator of a generative adversarial network (GAN; Goodfellow et al. 2014; Section 20.10.4 of Goodfellow et al. 2016). The discriminator learns high-level features of images, which are more closely related to human perception of image similarity than are pixelwise measures. Stengel et al. (2020) thus called the adversarial loss a "perceptual loss," and their Figures 3 and 7 show that it leads to more realistic upsampled images than does the pixelwise MSE.

Besides perceptual loss functions, some work in computer science and remote sensing uses local spatial gradients to spatially enhance the loss function (*e.g.*, Eigen and Fergus 2015). This is most often done by using the SSIM index as the loss function (Zhao et al. 2016; Hammernik et al. 2017; Snell et al. 2017; Harder et al. 2020).

*c. Emphasis and scope of this study*

In Ebert-Uphoff et al. (2021), our group provided an introduction to custom loss functions in the geosciences, including the idea of using the FSS as a loss function. This idea was tested for the first time in Lagerquist et al. (2021a, henceforth L21a), for the problem of forecasting convection at 0–2-hour lead times. Specifically, L21a used the 9-by-9-pixel FSS as a loss function and demonstrated that U-nets trained with the 9-by-9 FSS outperform those trained with a pixelwise loss function. However, L21a explored only this one SELF, whereas in this paper we focus entirely on exploring many SELFs in depth. Specifically, from the four types of SV methods discussed in Gilleland et al. (2009), we incorporate two in loss functions: neighbourhood and scale-separation methods. We turn verification scores common in atmospheric science, such as the Brier score and FSS, into both neighbourhood and scale-separation loss functions. The latter use spectral filtering, via Fourier or wavelet decomposition, to separate scales[1]. Our prediction task is the same as in L21a, but we focus only on the 1-hour lead time.

We address four scientific questions in this study. (1) What are the options for neighbourhood and scale-separation loss functions? (2) How can these loss functions be implemented? (3) How do these loss functions influence the results? (4) For a given geoscience application, can we use expert knowledge to determine which loss functions should (not) be used? To our knowledge, this paper is the first in-depth exploration of SELFs in the geosciences. We have included code for all SELFs (see "Data availability statement"), using the Keras and TensorFlow libraries in Python (Chollet et al. 2015).

## 2. Sample application

This section briefly introduces our sample application: the prediction task (1-hour forecasting of convection), input data, and an explanation of the U-net NN architecture. See L21a for further details.

As predictors for convection, we use a time series of brightness-temperature maps from the Himawari-8 satellite, with seven spectral bands (Figures 1a-g) and three lag times. The seven spectral bands – 6.25, 6.95, 7.35, 8.60, 10.45, 11.20, and 13.30 $\mu$m – are all in the infrared portion of the electromagnetic spectrum, so our U-nets can be used during both day and night. The three lag times are 0, 20, and 40 minutes before the forecast-issue time $t_0$. As targets (treated as correct answers during U-net-training), we use a convection mask at $t_0 + 1$ hour. The mask is produced by applying a modified version of Storm-labeling in 3 Dimensions (SL3D; Starzec et al. 2017), an echo-classification algorithm, to radar data from Taiwan (Figure 1h). The mask contains 1 at pixels with a thunderstorm and 0 elsewhere (Figure 1i). Both the satellite and radar data are provided by the Taiwan Central Weather Bureau on a 0.0125° grid. We verify the U-nets only at pixels within 100 km of the nearest radar (Figure 1i), deemed to have adequate coverage for detecting convection.

---

[1]However, for reasons that will become clear in Section 4c, it is better to perform spectral filtering outside the loss function.

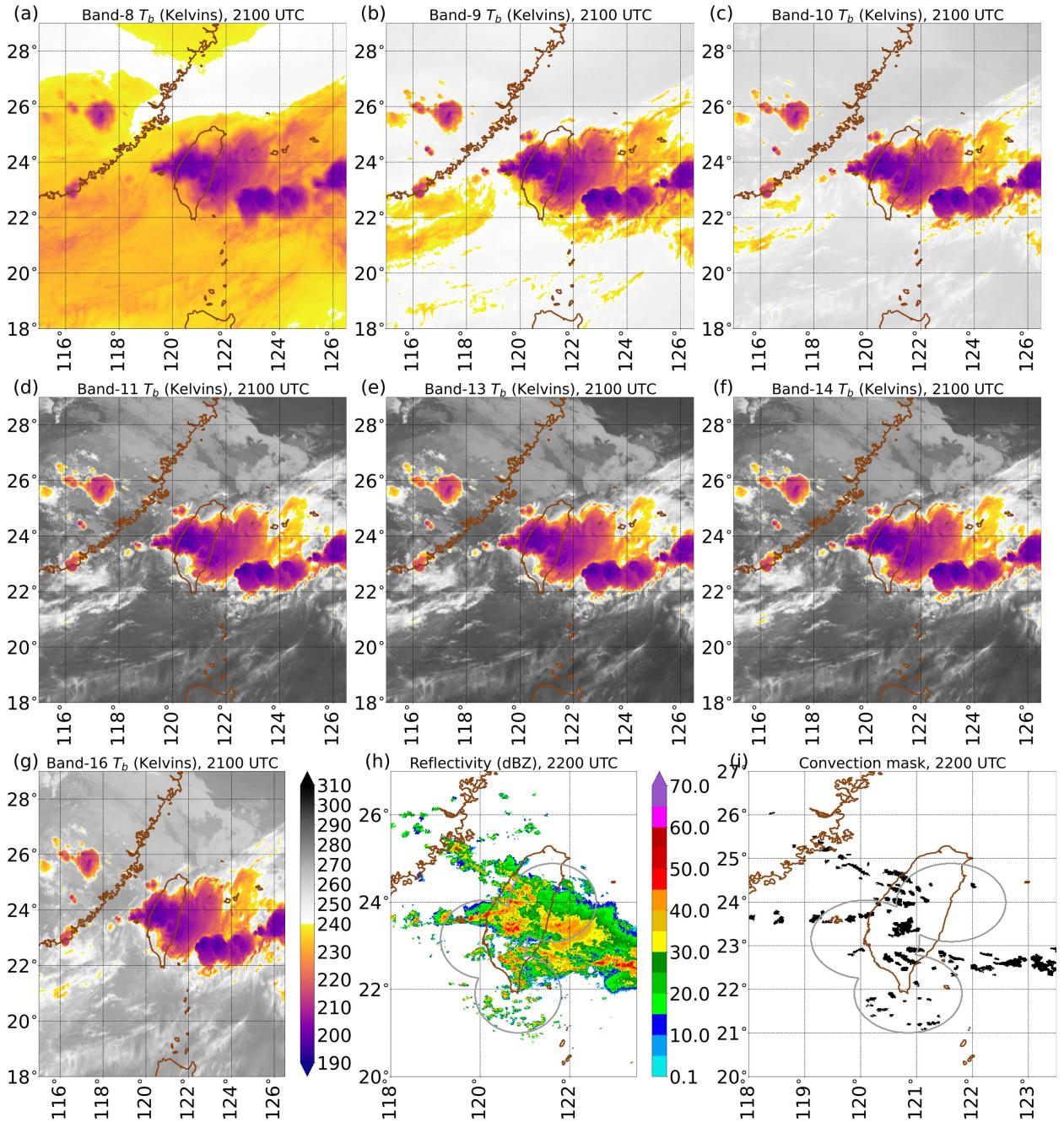

Figure 1: Example of input data for U-net, featuring predictors valid at 2100 UTC 2 Jun 2017 and targets valid one hour later, at 2200 UTC. [a-g] Brightness temperature (Kelvins) in each spectral band, used as predictors. All panels a-g use the colour bar next to panel g. [h] Composite (column-maximum) radar reflectivity. [i] Convection mask. The black dots are pixels with convection, according to the SL3D algorithm used to create targets. Grey circles in panels h-i show the 100-km range ring around all but the northernmost radar. Only pixels inside these range rings are used to train and verify the U-nets. The northernmost radar is omitted from training and verification, due to data-quality issues discussed in L21a.

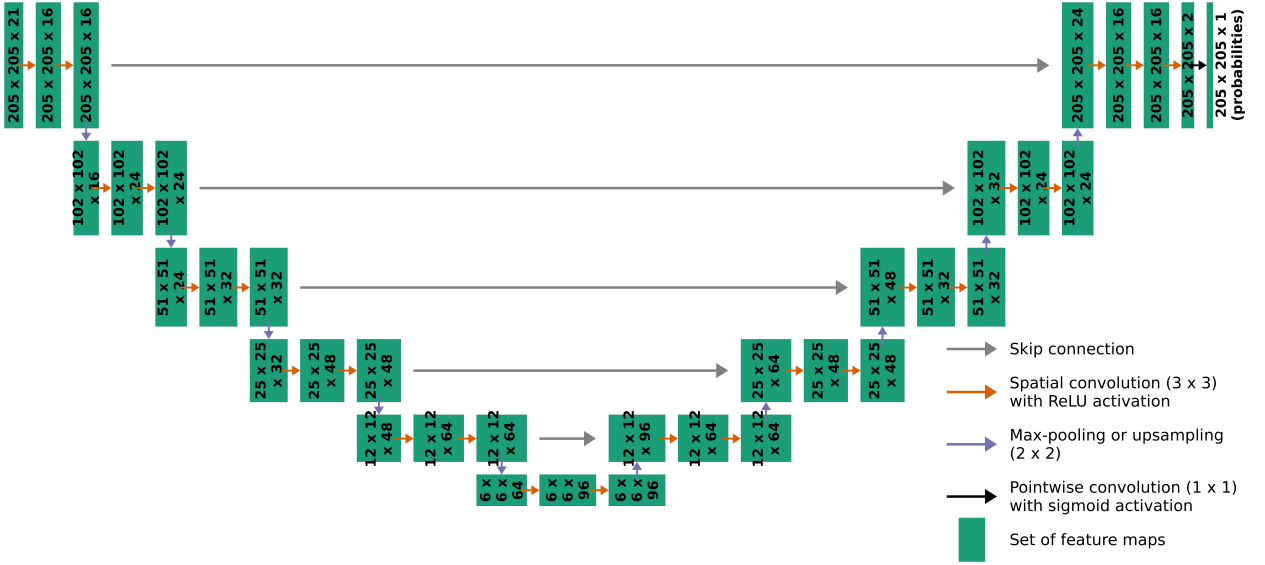

Figure 2: U-net architecture. The left (right) side is the downsampling (upsampling) side. In each set of feature maps, the numbers are $N_{\text{rows}} \times N_{\text{columns}} \times N_{\text{channels}}$. The 21 input channels at the top left are the predictor variables, *i.e.*, brightness temperatures from 7 spectral bands at 3 lag times. Spatial-convolution filters have dimensions of 3 rows × 3 columns; pixelwise-convolution filters have dimensions of 1 × 1; pooling and upsampling windows have dimensions of 2 × 2.

Our chosen NN architecture is the U-net, which excels at gridded prediction (Ronneberger et al. 2015). A U-net contains four components (Figure 2): convolutional layers, pooling (downsampling) layers, upsampling layers, and skip connections. The convolutional layers detect spatial and multivariate features (here, spatial features involving many spectral bands and lag times); all other components allow different convolutional layers to detect features at different scales. Inputs to the first layer are raw predictors (here, brightness temperatures), and inputs to all other layers are transformed versions of the raw predictors, called feature maps. Each convolutional layer is followed by a non-linear activation function; without these, the U-net would learn only linear relationships. Our chosen activation function is the leaky rectified linear unit (ReLU; Maas et al. 2013). Each pooling layer downsamples the feature maps to a lower spatial resolution, here using a 2-by-2 maximum filter. Hence, on the left (downsampling) side of Figure 2, grid spacing varies from $0.0125°$ at the top to $0.4°$ at the bottom. As the spatial resolution decreases, the number of feature maps ("channels") increases, to offset the loss of spatial information. Each upsampling layer upsamples the feature maps to a higher spatial resolution, using interpolation followed by convolution. Hence, on the right (upsampling) side of Figure 2, grid spacing varies from $0.4°$ at the bottom to $0.0125°$ at the top. As the spatial resolution increases, the number of channels decreases, terminating here with one channel (convection probability). Skip connections carry high-resolution information from the downsampling side of the U-net directly to the upsampling side, which is crucial because upsampling alone is not sufficient to recover the spatial information lost during downsampling. Skip connections include convolutional layers, as explained in L21a. In all convolutional layers (on the downsampling side, on the upsampling side, and in the skip connections), weights in the convolutional kernels are learned during training. The final set of weights is that which minimizes the loss function. For more details on the U-net architecture, including the inner workings of all components, see Section 3b of L21a.

As mentioned above, we verify the U-nets only at pixels $<$ 100 km from the nearest radar. Thus, at pixels $>$ 100 km from the nearest radar, there is no target value (*i.e.*, neither 0 nor 1 in the convection mask). To handle missing target values, we train each U-net with 205-by-205 radar-centered patches of the full 881-by-921 grid. At inference time, to apply a trained U-net to the full grid, we slide the 205-by-205 window around the full grid. For more details on these training and inference methods, see Sections 4a-b of L21a.

## 3. Neighbourhood loss functions

This section defines the neighbourhood loss functions used and discusses how we implement them in code.

*a. Definition*

A neighbourhood loss function involves a mean or maximum filter, taken over a square neighbourhood of pixels. The neighbourhood dimensions are always odd (*e.g.*, $3 \times 3$, $5 \times 5$), so that the center aligns with one pixel. A key motivation for neighbourhood loss functions is that they avoid the double-penalty problem (discussed in Section 1), because they always expand convective regions (Figure 3), thus increasing overlap between the predictions and observations. Compared to scale-separation methods, the main advantages of neighbourhood methods are ease of implementation (*c.f.* Sections 3b and 4c) and transparency. To the second point, it is easy to guess how a neighbourhood filter will change the original data (Figure 3), not so for the filters involved in scale separation (Figure 4).

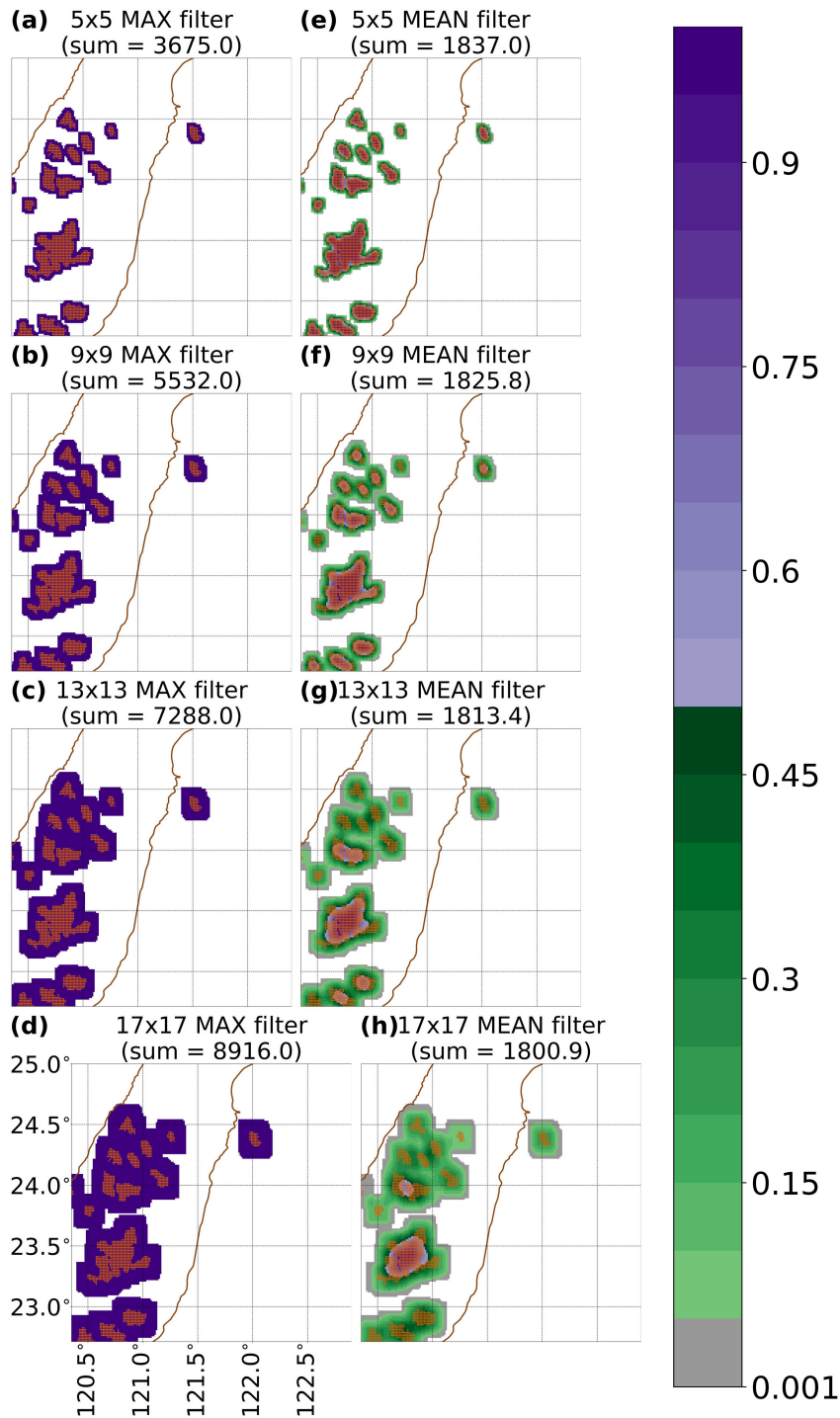

Figure 3: Effect of neighbourhood filtering on target fields. Orange dots on top show the target field before filtering (convective pixels are marked with an orange dot, while non-convective pixels have no orange dot); the colour map shows the target field after filtering. Before filtering, all target values are exactly 0 or 1; after filtering, target values range continuously from 0 to 1. The sum in each panel title is the sum over all pixels after filtering. This figure shows 4 of the 8 neighbourhood sizes used in the experiment (see Table 3 for full list). [a-d] Target fields, valid at 2200 UTC 2 Jun 2017, after maximum filter, used for all scores except FSS (Table 1). [e-h] Same but with mean filter, used for FSS.

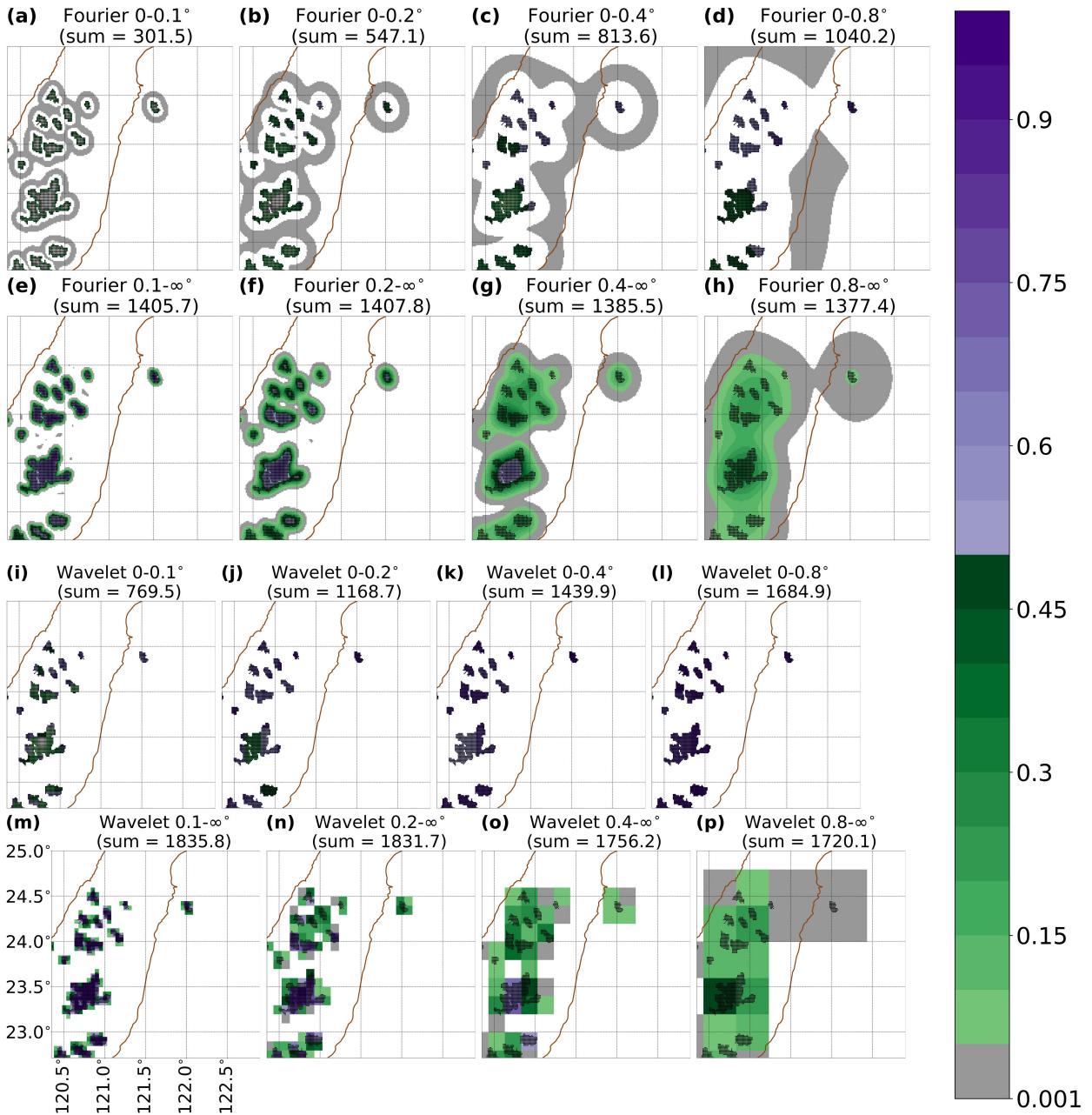

Figure 4: Effect of spectral filtering on target fields. Formatting is as in Figure 3, except that convective pixels are marked with a black dot rather than an orange dot. (We use black dots henceforth in the paper; we used orange dots in Figure 3 because black did not contrast enough with the large areas of dark purple.) This figure shows 8 of the 16 wavelength bands used in the experiment (see Table 3 for full list). [a-h] Target fields, valid at 2200 UTC 2 Jun 2017, after spectral filter implemented via Fourier decomposition. [i-p] Same but with wavelet decomposition.

Although neighbourhood loss functions are easy to code and appropriate for many atmospheric-science problems, only a few studies have done this and they use only the FSS (Lagerquist et al.

2021a; Earnest et al. 2022; Justin et al. 2022). We go beyond the FSS and turn five other common verification scores into neighbourhood loss functions: the Brier score, intersection over union (IOU), Dice coefficient, critical success index (CSI), and cross-entropy. Table 1 defines the traditional and neighbourhood-based version of all six scores. The traditional version is pixelwise for all scores except the FSS, which is already neighbourhood-based[2]. Variables in Table 1 are defined below, along with an intuitive definition of each score. For positively oriented scores $s$ (where higher values are better; all except the Brier score and cross-entropy), we use $1 - s$ as the loss function, since loss functions are negatively oriented. We do the same for scale-separation loss functions (Section 4).

In the traditional Brier score, $G$ is the number of pixels; $p_g$ is the predicted event probability at pixel $g$; and $y_g$ is the observation (0 or 1) at $g$. The traditional Brier score is the MSE adapted for binary classification. In the neighbourhood Brier score, $r$ is the neighbourhood half-width and $y_g^{\max}(r)$ is the maximum observation in the neighbourhood (Figure 5). Thus, $y_g^{\max}(r)$ is 1 if the event occurs anywhere in the neighbourhood and 0 otherwise.

In the FSS, $\overline{p_g}(r)$ is the mean probability, while $\overline{y_g}(r)$ is the mean observation, in a neighbourhood of half-width $r$ centered at pixel $g$. The numerator is the actual sum of squared errors (SSE) between mean-filtered predictions and observations, while the denominator is the reference SSE – the maximum SSE possible given the two fields, which occurs if they are completely misaligned.

Note that the FSS traditionally involves binary predictions (0s and 1s) but a U-net, like most machine-learning models for classification, outputs probabilities ranging continuously from $[0, 1]$. These probabilities *could* be transformed to binary predictions by discretization, *i.e.*, thresholding the probabilities. However, the best threshold is not clear and is typically not the intuitive guess of 0.5. One *could* find the best probability threshold (*i.e.*, that which minimizes the loss function) every time the loss function is evaluated, but this operation would be computationally expensive and would involve the non-differentiable minimum operator[3]. Thus, for all scores other than the Brier score and cross-entropy (which are already probabilistic), we replace binary predictions in the traditional definition with probabilities. We do the same for scale-separation loss functions (Section

---

[2]The FSS can be turned into a pixelwise score by using a single pixel, *i.e.*, a 1-by-1 neighbourhood. In this case, the equation in Table 1 becomes $1 - \dfrac{\sum\limits_{g=1}^{G}(p_g - y_g)^2}{\sum\limits_{g=1}^{G}(p_g^2 + y_g^2)}$.

[3]All loss functions must be differentiable, as discussed further in Section 4c.

Table 1: Verification scores used in both neighbourhood and scale-separation loss functions. Each score is defined for one time step; to obtain the score for multiple time steps, compute the score for each time step individually and then average. For a scale-separation loss function, the traditional (pixelwise) definition of the score is applied to the spectrally filtered predictions and observations. The range of possible values is $[0, \infty)$ for cross-entropy and $[0, 1]$ for all other scores. Variables in the equations, as well as an intuitive definition of each score, are provided in the main text.

| Score | Traditional Definition | Neighbourhood-based Definition | Optimal Value |
|---|---|---|---|
| Brier score | $\frac{1}{G} \sum_{g=1}^{G} (p_g - y_g)^2$ | $\frac{1}{G} \sum_{g=1}^{G} \left[ p_g - y_g^{\max}(r) \right]^2$ | 0 |
| FSS | N/A | $1 - \dfrac{\sum_{g=1}^{G} \left[ \overline{p_g}(r) - \overline{y_g}(r) \right]^2}{\sum_{g=1}^{G} \left[ \overline{p_g}^2(r) + \overline{y_g}^2(r) \right]}$ | 1 |
| IOU | $\dfrac{\sum_{g=1}^{G} p_g y_g}{\sum_{g=1}^{G} \max(p_g, y_g)}$ | $\dfrac{\sum_{g=1}^{G} p_g y_g^{\max}(r)}{\sum_{g=1}^{G} \max\{p_g, y_g^{\max}(r)\}}$ | 1 |
| Dice coefficient | $\dfrac{\sum_{g=1}^{G} p_g y_g + \sum_{g=1}^{G} (1-p_g)(1-y_g)}{G}$ | $\dfrac{\sum_{g=1}^{G} p_g y_g^{\max}(r) + \sum_{g=1}^{G} \left[ 1-p_g \right]\left[ 1-y_g^{\max}(r) \right]}{G}$ | 1 |
| CSI | $\dfrac{a}{a+b+c}$ | $\text{CSI}^{-1}(r) = \text{POD}^{-1}(r) + \text{SR}^{-1}(r) - 1$<br>$\text{POD}(r) = \dfrac{a_{\text{obs}}(r)}{a_{\text{obs}}(r) + c(r)}$<br>$\text{SR}(r) = \dfrac{a_{\text{pred}}(r)}{a_{\text{pred}}(r) + b(r)}$ | 1 |
| Cross-entropy | $\dfrac{\sum_{g=1}^{G} \left[ y_g \log_2(p_g) + (1-y_g)\log_2(1-p_g) \right]}{G}$ | $\dfrac{\sum_{g=1}^{G} \left[ y_g^{\max}(r) \log_2(p_g) + (1-y_g^{\max}(r))\log_2(1-p_g) \right]}{G}$ | 0 |

4). This changes the physical meaning of the FSS, which was originally meant to compare the forecast and observed event coverage (here, percentage of convective pixels in the neighbourhood). Here, we compare the observed event coverage to the spatial sum of forecast event probabilities, rather than the forecast event coverage. In addition to the technical challenges that come with thresholding, we speculate that using raw probabilities in the loss functions allows the NNs to achieve better performance. The magnitudes of the NN-forecast probabilities contain information, and thresholding destroys much of this information.

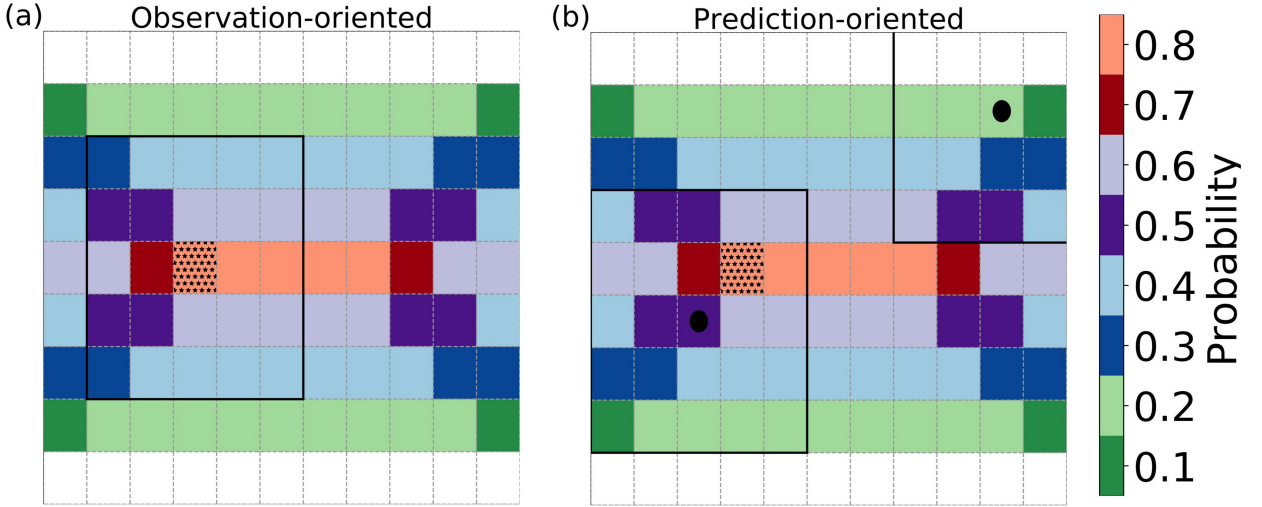

Figure 5: Model verification with a 5-by-5 neighbourhood (*i.e.*, half-width of 2 pixels). The colour map shows predicted probabilities; convection is observed at the one grid cell hatched with asterisks (*) and nowhere else. [a] Observation-oriented evaluation, where the central pixel is the one with observed convection. The maximum probability in the neighbourhood is 0.8, so this situation contributes 0.8 to $a_{\text{obs}}$ (observation-oriented true positives) and $1 - 0.8 = 0.2$ to $c$ (false negatives). [b] Prediction-oriented verification, shown for two central pixels (marked with circles). For the bottom-left pixel (probability of 0.5), convection is observed in the neighbourhood, so this situation contributes 0.5 to $a_{\text{pred}}$ (prediction-oriented true positives) and $1 - 0.5 = 0.5$ to $b$ (false positives). For the top-right pixel (probability of 0.2), convection is *not* observed in the neighbourhood, so this situation contributes 0.2 to $b$ (false positives). Because there is no convection in the neighbourhood, this situation contributes nothing to $a_{\text{pred}}$ (prediction-oriented true positives).

In the traditional IOU, $\max(p_g, y_g)$ is the maximum between the prediction and observation at pixel $g$. The numerator is the intersection between the two fields (where the event both occurs and is predicted), and the denominator is the union (where the event either occurs or is predicted). In the neighbourhood IOU, $y_g^{\max}(r)$ is defined as in the neighbourhood Brier score. Note that we use only the positive-class IOU: for class 1 (convection), rather than class 0 (no convection).

The Dice coefficient is similar to the IOU; the main difference is that the denominator is the area of the full domain, rather than the intersection between predicted and observed events. For spatially rare events, if defined only for the positive class, the Dice coefficient is typically very small, because the full domain is much larger than either the predicted or observed event areas. Thus, we use the all-class Dice coefficient. In the traditional Dice coefficient, $1 - p_g$ is the negative-class probability and $1 - y_g$ is the negative-class label (0 if the event occurs, 1 if it does not). The definition of $1 - y_g^{\max}(r)$ is analogous but with the convection labels maximum-filtered over a neighbourhood with half-width $r$.

The CSI is based on the contingency table for binary classification, which traditionally has four elements: the number of true positives ($a$), false positives ($b$), false negatives ($c$), and true negatives ($d$). However, in the neighbourhood setting, there are two types of true positives: prediction-oriented and observation-oriented (Figure 5). The National Weather Service verifies tornado warnings in a similar setting (Brooks 2004), where the contingency table has the following elements: number of prediction-oriented true positives ($a_{\text{pred}}$), observation-oriented true positives ($a_{\text{obs}}$), false positives ($b$), and false negatives ($c$). In the neighbourhood equation for CSI, POD is the probability of detection and SR is the success ratio. Although POD and SR are defined differently in the traditional and neighbourhood settings, $\text{CSI}^{-1}(r) = \text{POD}^{-1}(r) + \text{SR}^{-1}(r) - 1$ holds in both settings (Brooks 2004; Roebber 2009). Because we do not discretize probabilities from the U-nets, we use a probabilistic form of the contingency table (Figure 5).

Cross-entropy (often called "logarithmic loss" or "log loss") is a commonly used loss function in machine learning, not limited to atmospheric-science applications. For a perfect model, $p_g = 1$ wherever $y_g = 1$, causing the numerator to be $1 \log_2(1) + 0 \log_2(0) = 1(0) + 0(-\infty) = 0$. Also for a perfect model, $p_g = 0$ wherever $y_g = 0$, causing the numerator to be $0 \log_2(0) + 1 \log_2(1) = 0(-\infty) + 1(0) = 0$. Hence, the perfect cross-entropy is zero.

## b. Implementation

Neighbourhood loss functions are easy to implement in NN libraries, because the mean and maximum filters used can be achieved via convolution and pooling respectively, which are predefined NN operations and are differentiable. Hence, we put neighbourhood filters directly in the loss function, as illustrated in Figure 6. See "Data availability statement" for a link to the code.

## 4. Scale-separation loss functions

Compared to neighbourhood methods, the main advantage of scale-separation methods is their ability to focus on a specified *range* of scales, with an upper and lower bound. Neighbourhood filters always preserve (degrade) information at the larger (smaller) scales, but the spectral filters used in scale separation preserve (degrade) information at scales inside (outside) the desired range.

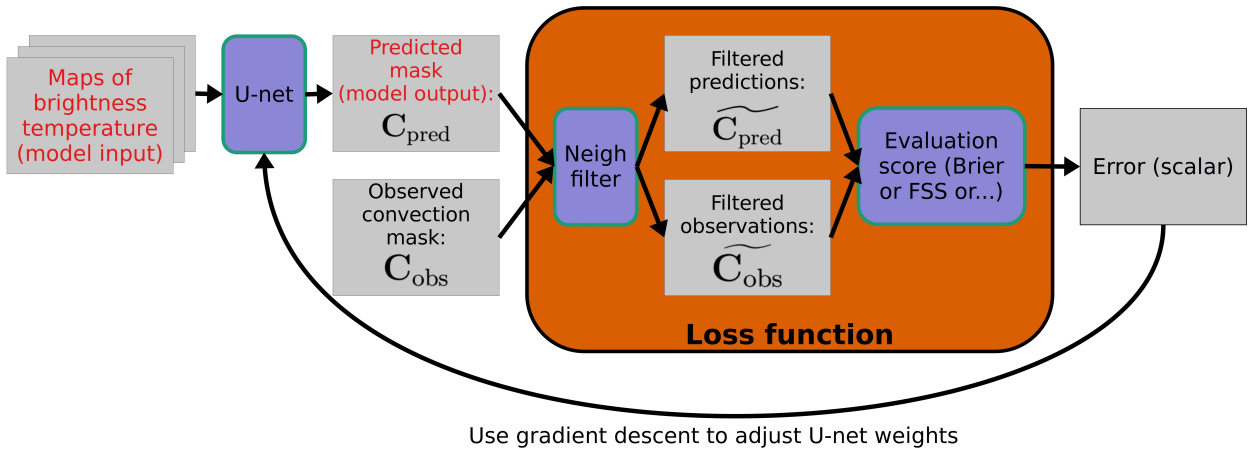

Figure 6: Implementation of neighbourhood filter inside a U-net. The neighbourhood filter is embedded inside the loss function, where it is applied to both the predictions and the observations.

We explore both Fourier and wavelet decomposition for scale separation, because (a) they are the most common methods and (b) they have quite different properties, as discussed later in this section, and it is not obvious how these properties will influence what an NN learns.

The remainder of this section begins with a discussion of previous work combining Fourier and wavelet decomposition with NNs, then defines the scale-separation loss functions that we use, then discusses how we implement them in code.

*a. Previous work combining Fourier and wavelet decomposition with neural networks*

To our knowledge we are the first to explore using Fourier or wavelet decomposition in loss functions, but in general, the idea of using Fourier and wavelet decomposition in NNs is not new.

Many studies have used Fourier decomposition to accelerate the training of convolutional neural networks (CNN), leveraging the fact that convolution in the spatial domain is equivalent to elementwise multiplication in the Fourier domain (Mathieu et al. 2014; Rippel et al. 2015; Pratt et al. 2017). Also, Lee-Thorp et al. (2021) replaced the self-attention sublayer (which contains many learned parameters) in natural-language-processing models with a Fourier transform (which contains no learned parameters), achieving comparable accuracy in much less computing time. Fujieda et al. (2017) replaced pooling layers in CNNs (which keep only the low-frequency part of an image) with wavelet-decomposition layers (which keep both the high- and low-frequency parts), allowing for better texture classification. Li et al. (2020a) replaced spatial pooling in CNNs with spectral truncation (filtering of wavelet coefficients), which is better at separating the signal and noise (*i.e.*, desired and undesired scales). Lastly, some works have used decomposition methods to transform predictors from spatial images to spectral (Fourier or wavelet) coefficients, then trained NNs with only the coefficients, achieving similar accuracy to training with spatial images but in much less computing time (*e.g.*, Kiruluta 2017). Other examples of training NNs with spectral coefficients include at least two geoscience applications: predicting daily precipitation (Partal et al. 2015) and daily river discharge (Gürsoy and Engin 2019).

Of significant interest to atmospheric science is the recent development of the neural operator and Fourier neural operator (FNO), which are special types of NN architectures. Neural operators (Lu et al. 2019; Li et al. 2020b) can learn mappings between infinite-dimensional function spaces (*e.g.*, continuous fields), whereas traditional NNs can learn mappings only between finite-dimensional Euclidean spaces (*e.g.*, images with values at discrete grid points). This means that neural operators can learn an entire family of partial differential equations (PDE) in a grid-agnostic manner. However, neural operators are much slower than numerical PDE-solvers. Li et al. (2021) accelerated neural operators with Fourier transforms, resulting in FNOs. Li et al. (2021) demonstrated the impressive capabilities of FNOs by modeling the Navier-Stokes equation in turbulent flow. FNOs have since shown impressive skill in solving PDEs for physical problems, including atmospheric science. FNOs are a key component of FourCastNet (Pathak et al. 2022), arguably the most accurate machine-learning-based numerical weather prediction (NWP) model to date. Furthermore, FNOs are several orders of magnitude faster than traditional PDE-solvers, allowing larger NWP ensembles (*e.g.*, thousands of members) to be run with existing computational resources. Thus, FNOs will likely play a major role in future NWP-model development.

While FNOs use Fourier transforms solely to increase the speed of NN operations, our purpose for using Fourier transforms (and wavelet transform) is to implement image-filtering in spectral space – not to increase speed.

*b. Definition*

The traditional verification scores from Table 1, which we turn into neighbourhood loss functions (Section 3), we also turn into scale-separation loss functions. For all scores except the FSS, the equation used for the scale-separation loss function is the traditional equation in Table 1. For the FSS, whose traditional definition is already neighbourhood-based, we use a neighbourhood of 1 × 1 pixels. Thus, the equation for every scale-separation loss function is pixelwise. However, these loss functions are computed after applying a spectral filter to the observations, thus removing convection at undesired scales, and training the U-net to predict convection at the desired scales. Hence, the filtered observations at one pixel incorporate information from the rest of the spatial domain, so the scale-separation loss functions computed thereafter are spatially enhanced.

In addition to the six scores in Table 1, we turn the Heidke score, Peirce score, and Gerrity score (Table 2) into scale-separation loss functions. We do not turn these scores into neighbourhood loss functions, because their calculations involve true negatives. The neighbourhood-based verification setting involves two types of true negatives – observation-oriented and prediction-oriented, like the two types of true positives – but unlike true positives, it is unclear from the literature how to use the two types of true negatives in computing verification scores. Because of this ambiguity and the fact in a rare-event setting most true negatives are trivial (*i.e.*, it easy to predict no-convection in most cases), we do not use the Heidke, Peirce, or Gerrity scores in neighbourhood loss functions.

The Heidke score is the number of correct predictions relative to a random model; the Peirce score is similar but for a random *and unbiased* model; and the Gerrity score is similar to the Heidke score, except that the Gerrity score is equitable (gives random and constant models a score of 0, indicating no skill) and does not reward conservative models (*i.e.*, those that err on the side of not predicting a rare event). Like the CSI, equations in Table 2 involve the contingency table, for which we use a probabilistic form rather than discretizing to create binary predictions. For example, consider a pixel with actual convection and a predicted convection probability of 0.8. This pixel contributes 0.8 to *a* (number of true positives) and 1 - 0.8 = 0.2 to *c* (number of false negatives). As another example, consider a pixel with a predicted probability of 0.8 but no actual convection. This pixel contributes 0.8 to *b* (number of false positives) and 1 - 0.8 = 0.2 to *d* (number of true negatives).

For each target field – *i.e.*, the observed convection mask at one time step – we use either Fourier or wavelet decomposition to implement the spectral filter. The two procedures are described schematically in Figures 7-8 and in full detail in Supplemental Section 1. The main advantage of Fourier decomposition over wavelet decomposition is richness of representation. In transforming data from the spatial domain to the wavelength domain, for a spatial dimension with $N$ pixels,

Table 2: Verification scores used only in scale-separation loss functions. *a* is the number of true positives; *b* is the number of false positives; *c* is the number of false negatives; *d* is the number of true negatives; and $N = a + b + c + d$ is the total number of examples. For each score, 0.0 indicates no skill and 1.0 is the optimal value. An intuitive definition of each score is provided in the main text.

| Score | Definition | Range |
|---|---|---|
| Heidke score | $\frac{a+d-N_{\text{random}}}{N-N_{\text{random}}}$, where $N_{\text{random}} = \frac{1}{N}[(a+b)(a+c) + (b+d)(c+d)]$ | $(-\infty, 1]$ |
| Peirce score | $\frac{a}{a+c} - \frac{b}{b+d}$ or $\frac{a}{a+c} + \frac{d}{b+d} - 1$ | $[-1, 1]$ |
| Gerrity score | $\frac{ar^{-1} + dr - b - c}{N}$, where $r$ = event ratio = $\frac{N_{\text{events}}}{N_{\text{non-events}}} = \frac{a+c}{b+d}$ | $[-1, 1]$ |

Fourier decomposition represents $N$ wavelengths while wavelet decomposition represents only $\log_2(N)$ wavelengths. The main advantage of wavelet decomposition is simplicity: it does not require windowing in the spatial domain (Step 2 of Fourier procedure in Supplemental Section 1), and filtering out coefficients in the wavelength domain is much simpler (*c.f.* Step 4 of Fourier procedure and Step 3 of wavelet procedure, both in Supplemental Section 1).

*c. Implementation*

Although neighbourhood loss functions are easy to implement (Section 3b), in general implementing custom loss functions comes with major challenges (Ebert-Uphoff et al. 2021). First, NNs are trained via gradient descent (Section 6.5 of Goodfellow et al. 2016), so the loss function must be differentiable with respect to all NN weights. Second, the loss function must execute very quickly, since NN-training is already computationally expensive. Third, little guidance exists on writing custom loss functions, so researchers often encounter pitfalls.

First, we consider the implementation of loss functions with spectral filters using Fourier decomposition. The fast Fourier transform (FFT) and its inverse are pre-defined in many NN libraries. For example, in TensorFlow, the relevant methods are `tensorflow.signal.fft2d` and `tensorflow.signal.ifft2d`; both are differentiable. Other steps in Fourier decomposition, such as windowing and filtering (details in Supplemental Section 1), can be cast as elementwise matrix multiplication, which is also differentiable. Thus, at first we implemented spectral filtering by putting Fourier decomposition in the loss function, as shown in Figure 9a. This is analogous to the schematic for neighbourhood loss functions (Figure 6), except that the neighbourhood filter is replaced by a spectral filter using Fourier decomposition. In this setup, both the predictions and observations are filtered before they are compared to each other, but the NN's final output is the unfiltered predictions.

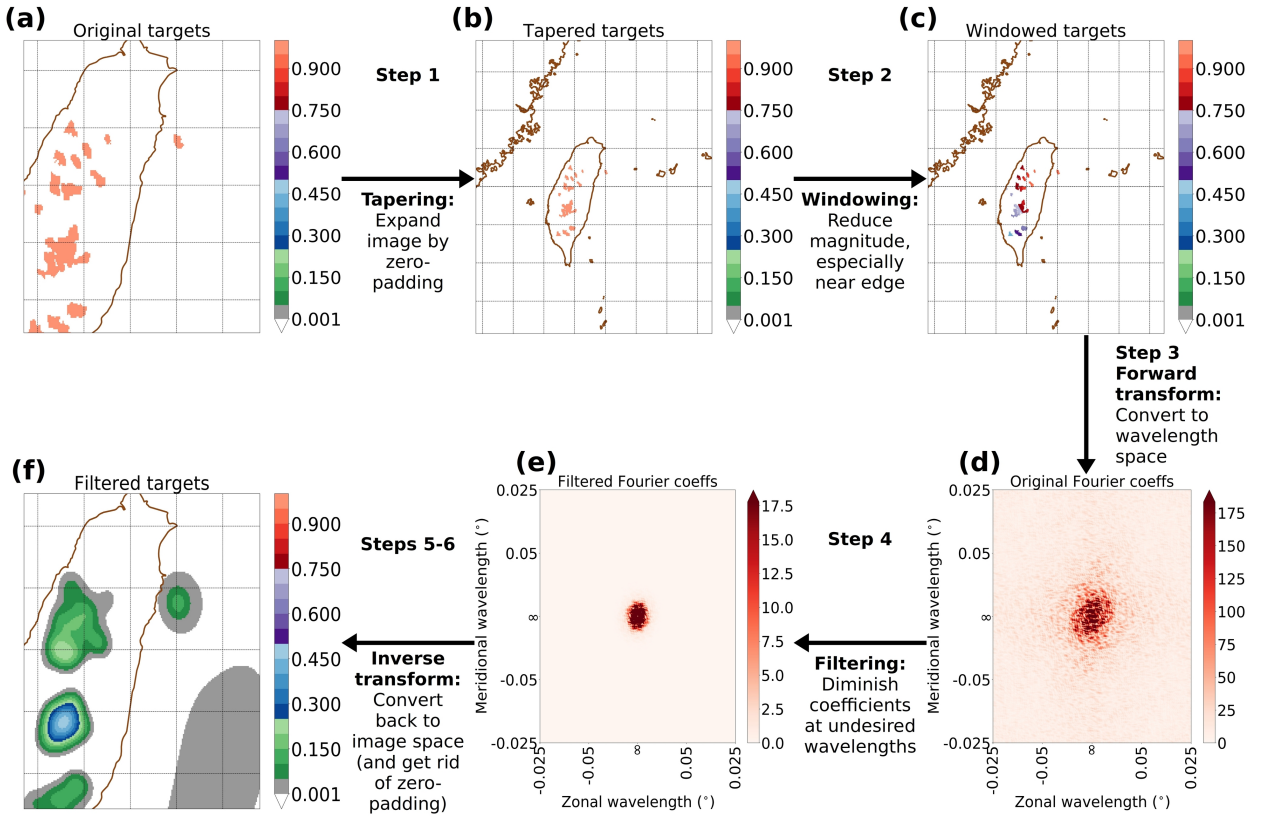

Figure 7: Implementation of spectral filter (in this case, allowing wavelengths of 0.5° to 2°) via Fourier decomposition. The original target field (*i.e.*, convection mask) contains only 0s and 1s, but after filtering, values range continuously from 0 to 1. For a detailed explanation of all steps in the procedure, see Supplemental Section 1. [a] Original convection mask, with dimensions of 205 × 205. [b] Tapered convection mask, with dimensions of 615 × 615. [c] Tapered convection mask after applying Blackman-Harris window (Supplemental Equation 1). [d] Fourier spectrum (*i.e.*, magnitude of each complex-valued coefficient), created by applying forward Fourier transform. [e] Fourier spectrum after applying Butterworth filter (Supplemental Equation 2). [f] Filtered convection mask, created by applying inverse Fourier transform and then removing zero-padding.

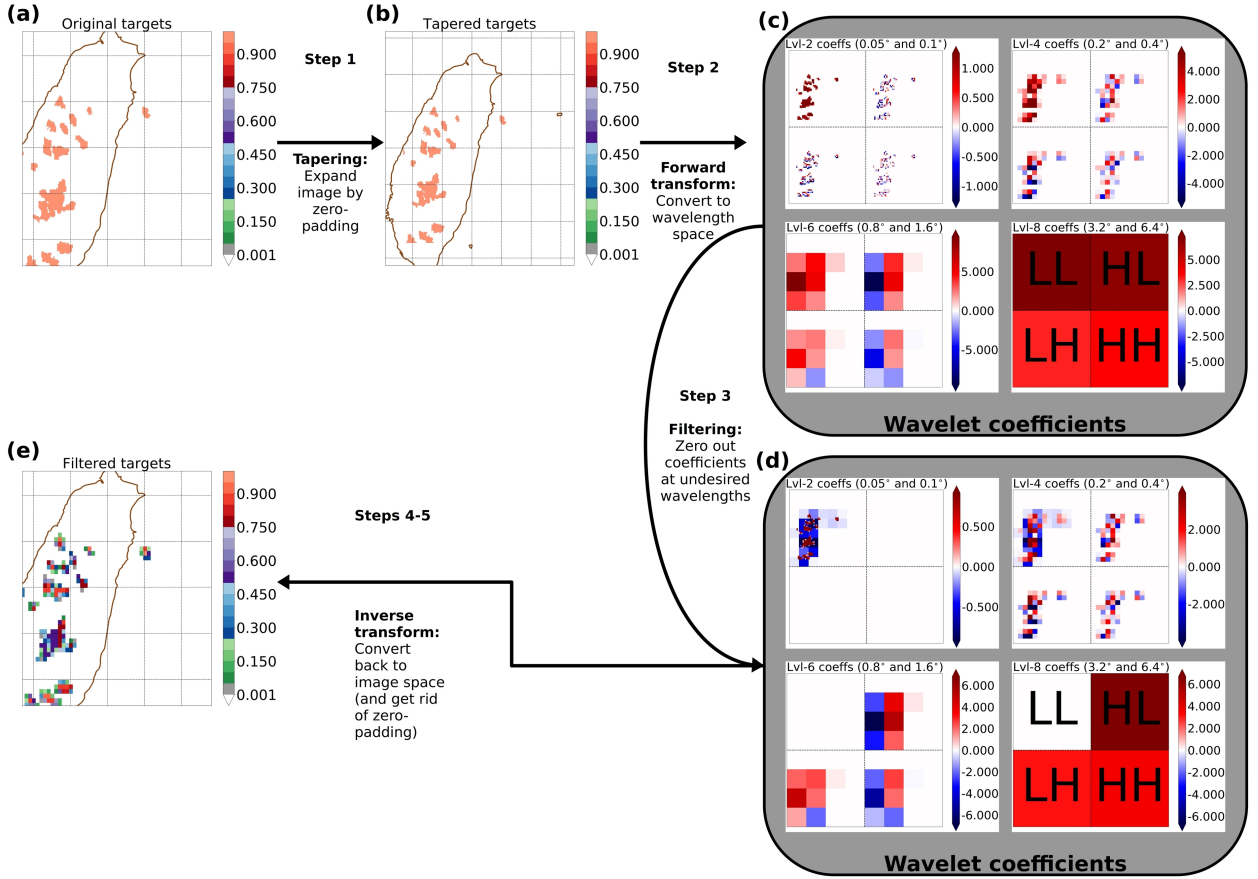

Figure 8: Implementation of spectral filter (in this case, allowing wavelengths of 0.1° to 0.4°) via wavelet decomposition. The original target field (*i.e.*, convection mask) contains only 0s and 1s, but after filtering, values range continuously from 0 to 1. For a detailed explanation of all steps in the procedure, see Supplemental Section 1. [a] Original field, with dimensions of 205 × 205. [b] Tapered field, with dimensions of 256 × 256. [c] Wavelet coefficients, created by applying forward wavelet transform. For brevity, only 4 of the 8 levels of decomposition are shown. Each subpanel here contains the LL, HL, LH, and HH coefficients (terms defined in main text), as marked in the bottom-right subpanel. At level 2, the grid size is 64 × 64; the smaller wavelength represented (the "H" part) is 0.05°; and the larger wavelength represented (the "L" part) is 0.1°. In general, at level $k$, the grid size is $M2^{-k} \times N2^{-k}$, where $M$ and $N$ are the grid size of the original data; the smaller wavelength represented is $\delta 2^k$, where $\delta$ is the grid spacing of the original data; and the larger wavelength represented is $\delta 2^{k+1}$. [d] Wavelet coefficients after filtering. [e] Filtered field, created by applying inverse wavelet transform and then removing zero-padding.

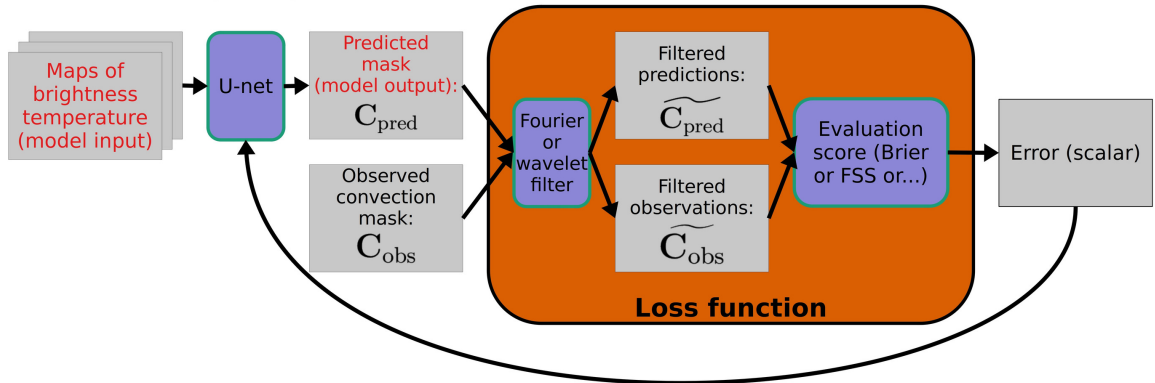
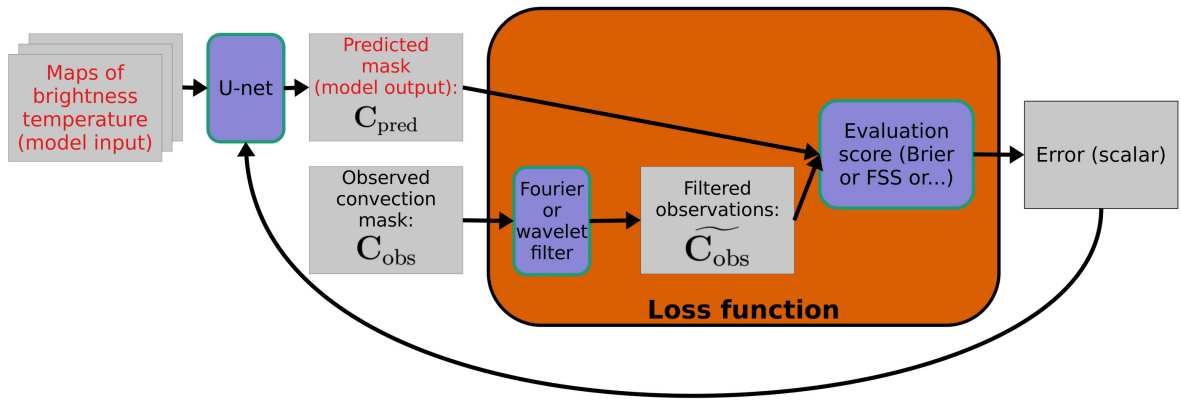
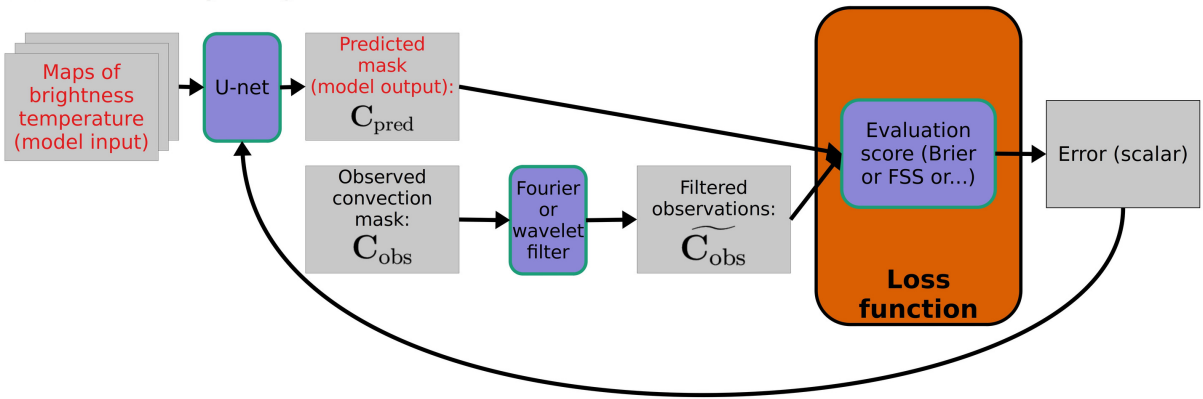

Figure 9: Three ways to implement a spectral filter inside a U-net. [a] Filter everything (the predictions and observations) directly inside the loss function. [b] Filter only the observations, doing so inside the loss function. [c] Filter only the observations, doing so outside the loss function (*i.e.*, as a pre-processing step). Methods b and c have the same effect, except that method c is less computationally expensive.

Experiments with this setup revealed a fundamental problem: the NN is free to produce arbitrary signals at the filtered wavelengths (*i.e.*, those censored out by the spectral filter) with no penalty. Specifically, the resulting U-nets produced convection probabilities that varied erratically in space and time and bore little resemblance to the observations. Why did the same problem not occur with neighbourhood loss functions? The spectral filters remove much more information from a spatial field than do the neighbourhood filters, which merely smooth or dilate over a small window (Figure 3), so neighbourhood loss functions give the NN less freedom. However, if the filter size is large enough, neighbourhood loss functions can lead to similar problems – see the discussion of "checkerboard patterns" in Section 6a. Henceforth, we will call this problem – giving the NN too much freedom at filtered scales – the *excessive-freedom problem*.

One possible solution to the excessive-freedom problem is using a loss function with two terms: $w_{\text{filt}} s_{\text{filt}} + w_{\text{unfilt}} s_{\text{unfilt}}$, where $s_{\text{filt}}$ is a score based on the filtered predictions and observations; $s_{\text{unfilt}}$ is the same but for unfiltered predictions and observations; and the $w_*$ are user-selected weights. However, this introduces more hyperparameters (the weights), which require careful tuning. Thus, we opted for a cleaner solution: filter the observations but not the predictions (Figure 9b). This setup forces the NN to produce signals only at the desired wavelengths (*i.e.*, those not removed by filtering), thus eliminating the excessive-freedom problem. However, if the spectral filter is applied only to the observations, it is completely equivalent – and more computationally efficient[4] – to apply the filter outside the loss function, as shown in Figure 9c. This is the setup we actually use for spectral filtering via Fourier decomposition.

To implement loss functions with spectral filters using wavelet decomposition, we went through a similar thought process. For wavelet decomposition, the forward and inverse transforms are pre-defined operations in the WaveTF library (Versaci 2021), which interfaces with TensorFlow, and are also differentiable. However, filtering the wavelet coefficients (details in Supplemental Section 1) involves iterating over decomposition levels (typically done with a `for` loop) and deciding which coefficients to zero out (typically done with an `if` statement), which are hard to write in a differentiable form. Owing to these challenges and the excessive-freedom problem that arises with spectral filtering, we decided to apply wavelet-based filters outside the loss function, as with Fourier-based filters (Figure 9c).

---

[4] During training, each observed field $\mathbf{C}_{\text{obs}}$ is read from disk many times. Filtering inside the loss function means filtering $\mathbf{C}_{\text{obs}}$ every time it is read from disk; filtering outside the loss function means filtering $\mathbf{C}_{\text{obs}}$ once and storing the filtered field on disk.

Table 3: Loss functions used in experiment. There are 48 neighbourhood loss functions, yielded by combining 6 verification scores × 8 filter sizes. There are 288 scale-separation loss functions, yielded by combining 9 verification scores × 16 scale ranges × 2 spectral-filtering methods (Fourier and wavelet decomposition). Hence, there is a total of 48 + 288 = 336 loss functions, resulting in 336 U-net models. The same 336 loss functions are used as verification metrics *post hoc*, *i.e.*, after training.

| Neighbourhood loss functions | |
| --- | --- |
| **Score** | **Neighbourhood half-width (pixels)** |
| Brier score | 0 (pixelwise), 1, 2, 3, 4, 6, 8, 12 |
| FSS | See above |
| IOU | See above |
| Dice coefficient | See above |
| CSI | See above |
| Cross-entropy | See above |
| **Scale-separation loss functions** | |
| **Score** | **Wavelength range (degrees)** |
| Brier score | $\leq 0.025$, 0.025-0.050, 0.050-0.100, 0.100-0.200, 0.200-0.400, 0.400-0.800, 0.800-1.600, $\geq 1.600$, $\leq 0.1$, $\leq 0.2$, $\leq 0.4$, $\leq 0.8$, $\geq 0.1$, $\geq 0.2$, $\geq 0.4$, $\geq 0.8$ |
| FSS | See above |
| IOU | See above |
| Dice coefficient | See above |
| CSI | See above |
| Cross-entropy | See above |
| Heidke score | See above |
| Peirce score | See above |
| Gerrity score | See above |

Note that the excessive-freedom problem can also occur for neighbourhood loss functions with larger neighbourhoods, as shown in Section 6a.

## 5. Experiment

**Combinations explored:** We train many U-nets, each with a different loss function (Table 3), to predict convection at 1-hour lead time. The goals of the experiment are to determine (1) the spatial characteristics of predictions (*i.e.*, probability maps) produced by training with each loss function, (2) which loss functions lead to the best performance overall, and (3) which loss functions lead to the best performance in certain situations. As shown in Table 3, the total number of loss functions – and hence the total number of U-nets trained – is 336.

**Training:** Following L21a and common practice in machine learning, we split the dataset into training (year 2016), validation (2017), and testing (2018). We use the training data to optimize model weights, the validation data to choose models for the final analysis (Section 6a-b), and the testing data for case studies in the final analysis (Section 6c). For other details of the training (optimizer, number of epochs, etc.), see L21a.

**Verification:** To verify the models after training (on the validation and testing data), we use a combination of visual (subjective) analysis and objective verification scores. In a study comparing the benefits of using different metrics as loss functions, it is non-trivial to decide which metric should be used for *post hoc* (after-training) verification. (Henceforth, we use the word "loss function" for a verification method used during training and "metric" for the same verification method used after training.) Thus, we use all 336 loss functions as metrics for *post hoc* verification.

## 6. Results

Section 6a presents the preliminary analysis, in which all models are compared on a case study in the validation data. We use the preliminary analysis to dismiss all but 120 models as inappropriate for the convection application. Section 6b presents the intermediate analysis, where we use the 336 verification metrics to compare the 120 models remaining. We use the intermediate analysis to select models for the final analysis, presented in Section 6c. In the final analysis, selected models are compared on several in-depth case studies in the testing data; we identify which models would be desirable for different risk thresholds and spatial scales of convection.

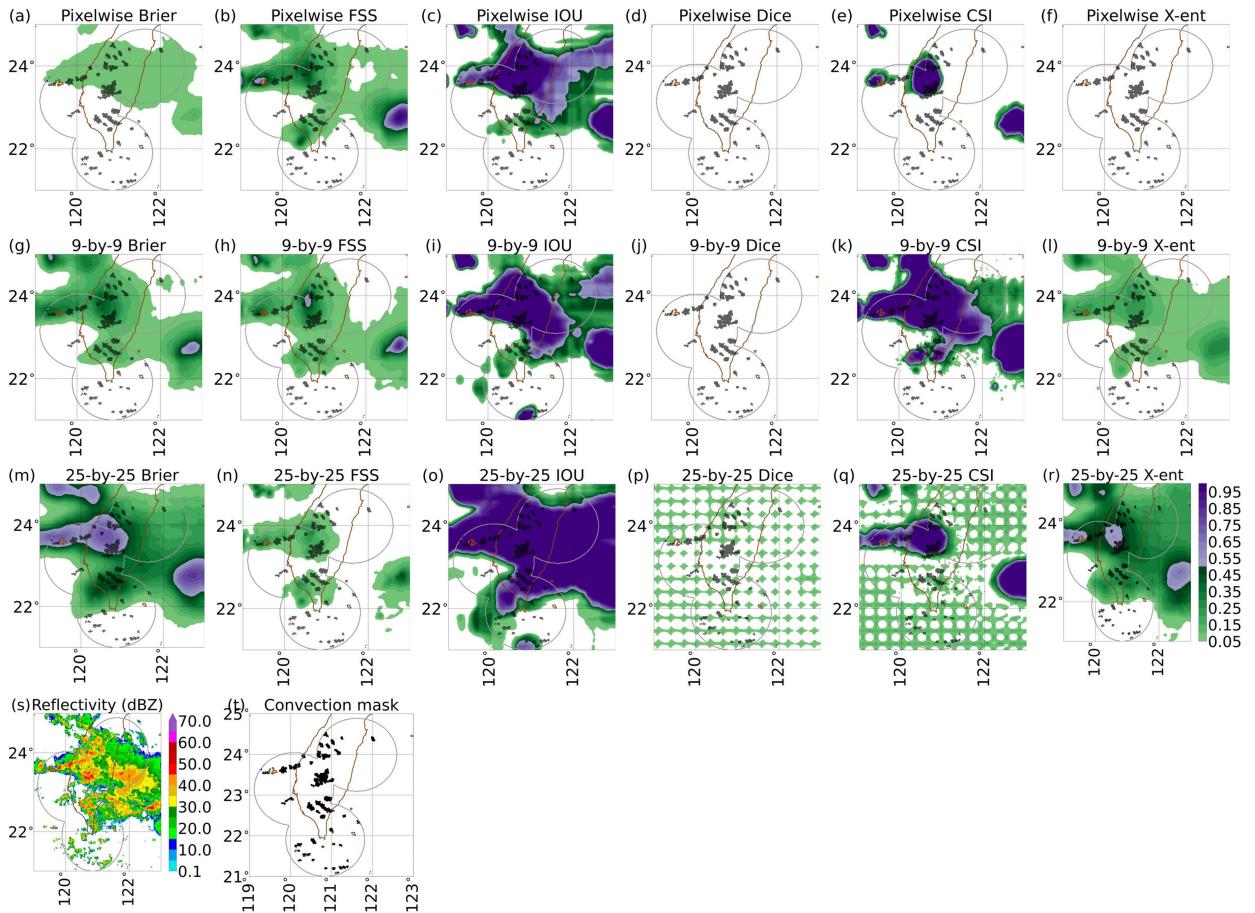

Figure 10: Predictions valid at 2200 UTC 2 Jun 2017, made by U-nets trained with different neighbourhood loss functions. Black dots show actual convection at 2200 UTC, according to the SL3D algorithm used to create targets. Black dots are not shown outside the 100-km range rings (grey circles), because targets (correct answers) here are unknown. [a-f] Convection probabilities forecast by U-nets trained to optimize scores with a 1-by-1 neighbourhood, *i.e.*, pixelwise scores. [g-l] Same but for U-nets trained to optimize scores with a 9-by-9 neighbourhood. [m-r] Same but for U-nets trained to optimize scores with a 25-by-25 neighbourhood. [s] Composite (column-maximum) radar reflectivity valid at 2200 UTC. [t] Convection mask valid at 2200 UTC.

*a. Preliminary analysis*

Figure 10 shows predictions made by models trained with neighbourhood loss functions, valid at 2200 UTC 2 Jun 2017, a randomly chosen time step in the validation data. Figure 11 is analogous but for scale-separation loss functions. Analyzing the various models' predictions yielded the following key results[5]:

---

[5]Although Figures 10-11 do not show all neighbourhood or spectral filters, we have confirmed that for the convection application the following observations generalize to all filters. We have also confirmed that the following observations generalize to other cases and are not specific to 2200 UTC 2 Jun 2017.

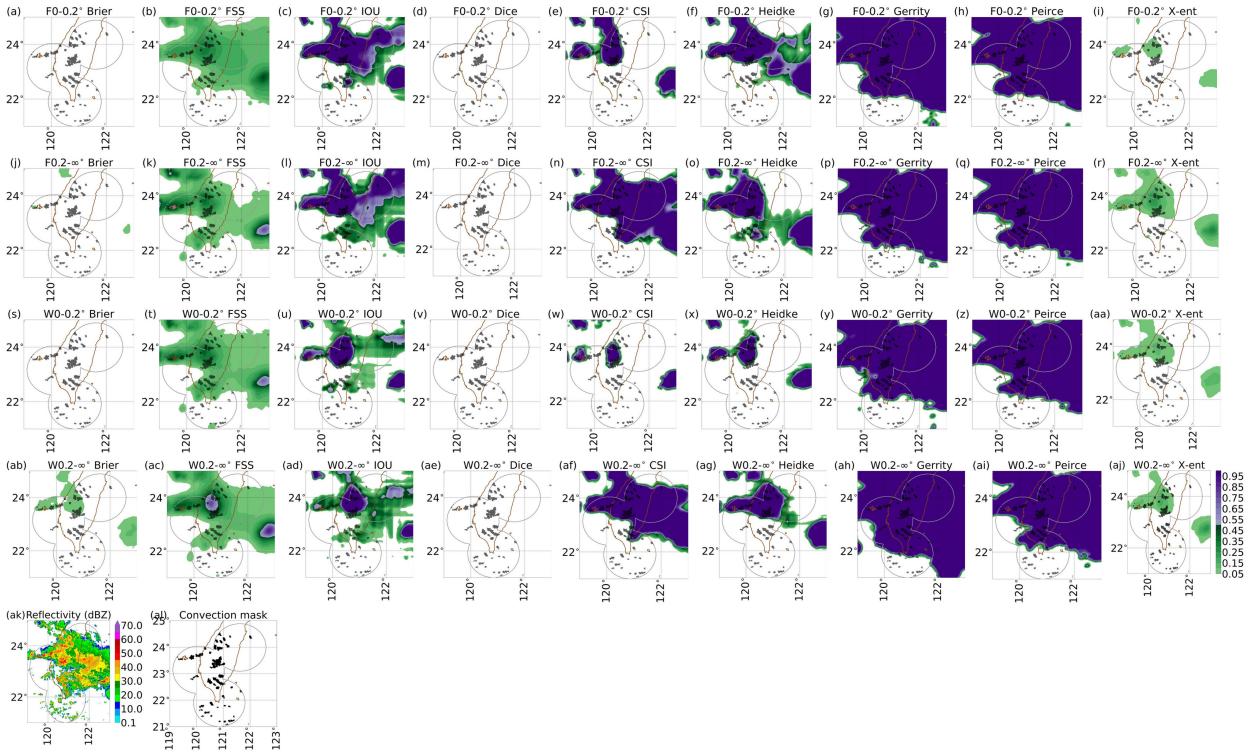

Figure 11: Predictions valid at 2200 UTC 2 Jun 2017, made by U-nets trained with different scale-separation loss functions. Formatting is explained in the caption of Figure 10. [a-i] Convection probabilities forecast by U-nets trained to optimize scores with filter "F0-0.2°," which uses Fourier decomposition to pass wavelengths of ≤ 0.2°. [j-r] Same but for U-nets trained to optimize scores with filter "F0.2-∞°," which uses Fourier decomposition to pass wavelengths ≥ 0.2°. [s-aa] Same but for U-nets trained to optimize scores with filter "W0-0.2°," which uses wavelet decomposition to pass wavelengths of ≤ 0.2°. [ab-aj] Same but for U-nets trained to optimize scores with filter "W0.2-∞°," which uses wavelet decomposition to pass wavelengths ≥ 0.2°. [ak] Composite (column-maximum) radar reflectivity valid at 2200 UTC. [al] Convection mask valid at 2200 UTC.

- Training with the Dice coefficient (fourth column of Figures 10 and 11) typically leads to very small probabilities (< 0.05). The Dice coefficient makes a poor loss function for rare events, because it rewards true negatives and true positives equally. Thus, for rare events, models can achieve a high Dice coefficient by always predicting negatives (no convection).

- Training with a 25-by-25 neighbourhood leads to curious spatial patterns, such as the checkerboard patterns in Figures 10p-q. These patterns are caused by the excessive-freedom problem (Section 4c), which arises for large neighbourhoods because the NN can produce erroneous small-scale patterns without penalty. When using neighbourhood filters, one should keep the filter size small enough to avoid this effect.

- Training with the CSI (fifth column of Figures 10 and 11) leads to large areas of very high probabilities ($\geq 0.95$), because the CSI rewards true positives but not true negatives. Interestingly, training with the CSI produces a very sharp model overall, with large areas of very low probabilities ($< 0.05$) as well, including many false negatives on the edge of the convective area.

- Training with the IOU (third column of Figures 10 and 11) and Heidke score (third-last column of Figure 11) causes similar problems.

- Training with the Gerrity or Peirce score (last two columns of Figure 11) also produces very high probabilities but, unlike the CSI and Heidke score, with fewer false negatives on the edge of the convective area[6,7].

- For models trained with the Brier score, FSS, or cross-entropy (first two columns and last column of Figures 10 and 11), the aforementioned issues with the other loss functions are not present, except that models trained with a scale-separation Brier score produce quite low probabilities ($< 0.05$ almost everywhere).

Based on the above observations, the rest of Section 6 will focus solely on models trained with the Brier score, FSS, or cross-entropy (120 of the 336 models).

*b. Intermediate analysis*

We use all 336 metrics for *post hoc* verification of the 120 remaining models – *i.e.*, those trained with the FSS, Brier score, or cross-entropy. The inputs to each metric are the unfiltered predictions and unfiltered observations ($\mathbf{C}_{\text{pred}}$ and $\mathbf{C}_{\text{obs}}$ in Figures 6 and 9). Most metrics (all except the six pixelwise ones listed in Table 1) involve a spatial filter, in which case $\mathbf{C}_{\text{pred}}$ and $\mathbf{C}_{\text{obs}}$ are eventually filtered *inside the metric*.

Applying all metrics to all models would require the verification of 120 models using 336 metrics, yielding $336 \times 120 = 40\,320$ individual scores. One way to simplify the intermediate analysis would be to consider only a subset of the 336 metrics, *i.e.*, to prioritize some metrics over others. However, the purpose of this study is to explore which of the metrics result in the most promising loss functions – and thus the most promising models – not which metrics make the most appropriate final evaluation criteria. The latter question is heavily dependent on the needs of the end user. Thus, we prefer to include all metrics in the evaluation of the models for this exploratory study. To avoid information overload (*i.e.*, considering 40 320 individual scores), we use the following procedure to generate for each model $\mathcal{M}$ a summary score for each spatial filter $\mathcal{F}$. This summary score can be interpreted as representing the general performance of $\mathcal{M}$ for the range of spatial scales highlighted by $\mathcal{F}$.

---

[6] The Gerrity score has a well known property of rewarding aggressive models (those that predict a rare event often), as documented in Chapter 4 of Jolliffe and Stephenson (2012).
[7] The Peirce score is $\frac{a}{a+c} + \frac{d}{b+d} - 1$, with variables defined in the caption of Table 2. $b+d$ is the number of non-events, and $a+c$ is the number of events, so $b+d \gg a+c$ for rare events. Thus, the second term has a much larger denominator, making the Peirce score more sensitive to the numerator of the first term (true positives) than that of the second term (true negatives).

The purpose of this evaluation procedure is to identify those models that stand out in their performance for a particular scale and to study their properties further in the case studies in Section 6c. Later studies with SELFs might take different approaches. In particular, if one knows for an application which metric best represents the goals of the end user, then one may use only that metric in model verification/selection. In contrast, here we want to explore any model that shows unusual abilities in any way for further study.

**Procedure to generate summary score for each spatial filter $\mathcal{F}$ and model $\mathcal{M}$:**

1. Compute all 336 metrics for $\mathcal{M}$.
2. Compute the rank on all 336 metrics for $\mathcal{M}$. For metrics involving a negatively oriented score – *i.e.*, the Brier score or cross-entropy – the model with the lowest (highest) value receives a rank of 1 (120). For metrics involving a positively oriented score – all except the Brier score or cross-entropy – the model with the lowest (highest) value receives a rank of 120 (1).

   **Purpose:** Converting the individual scores to ranks allows us to take meaningful averages across different metrics (see step 3).
3. For each spatial filter $\mathcal{F}$, find all metrics using $\mathcal{F}$ and average the ranks over said metrics. This is denoted as the model's summary score for $\mathcal{F}$.

   **Purpose:** This step provides a single summary score for each model and each spatial filter.

Just like for the loss functions used during training, the metrics used during *post hoc* verification employ 40 different spatial filters, listed in Table 3. Thus, the above procedure yields 40 summary scores for each model. Finally, for each spatial filter used in the metrics, we identify the model with the best summary score. Results are summarized below and in Figure 12.

- For 39 of 40 filters, the best model includes the FSS, not the Brier score or cross-entropy, in the loss function.
- For 20 of 40 filters, the best model is trained with pixelwise FSS.

- For 13 of 40 filters, the best model is trained with FSS after using wavelet decomposition to isolate wavelengths of 0.1-∞°. Henceforth, we call this W0.1-∞° FSS.
- For 1 of 40 filters, the best model is trained with F0.1-∞° FSS, which is the same as W0.1-∞° FSS but using Fourier decomposition.
- For 1 of 40 filters, the best model is trained with F0-0.4° FSS.
- For 1 of 40 filters, the best model is trained with cross-entropy using a 17-by-17 neighbourhood filter.
- For the other 4 filters, the best model is trained with a neighbourhood-based FSS. The most common neighbourhood size for the best model (occurring three times) is $3 \times 3$.

Of the models listed above, we wish to select a small number for the final analysis, which uses in-depth case studies to understand how the models behave in different situations. There are two clear winners in the list, the first being the model trained with pixelwise FSS – *i.e.*, no spatial enhancement at all. The second clear winner in the list is the model trained with W0.1-∞° FSS. The other scale-separation-based model in the list – trained with F0.1-∞° FSS – has the best summary score for only one filter, but we still choose this model for final analysis, for symmetry with the W0.1-∞° model. By comparing models trained with the same scales but different scale-separation methods (Fourier vs. wavelet), we can see how these two methods affect model behaviour. Furthermore, we choose the W0-0.1° and F0-0.1° models for final analysis. Although these models do not have the best summary score for any filter, the W0-0.1° and F0-0.1° filters used in their loss functions are complements to the filters used in the loss functions of the W0.1-∞° and F0.1-∞° models. This presents an opportunity to understand the differing effects of high- and low-pass filters on model behaviour. Lastly, we choose the model trained with 17-by-17 FSS for final analysis. This model also does not have the best summary score for any filter, but including the pixelwise (1-by-1) and 17-by-17 FSS[8] presents an opportunity to understand the differing effects of small and large neighbourhood filters on model behaviour.

*c. Final analysis*

In this section we analyze the properties of forecasts produced by the six final models: those trained with pixelwise, 17-by-17, W0.1-∞°, F0.1-∞°, W0-0.1°, and F0-0.1° FSS. We consider the pixelwise model a baseline, since it is not spatially enhanced, against which to compare the other models, which are spatially enhanced. Specifically, we analyze case studies including convection at many scales: discrete storm cells, multi-cell clusters, a quasi-linear convective system (QLCS), and a tropical cyclone. The goal is to compare characteristics of the models' predictions in a qualitative manner, so that we can subjectively evaluate the strengths and weaknesses of each loss function. For each time step shown in Figures 13-14, we have been careful to highlight differences in model performance that are representative of the entire corresponding day (Aug 23 and Jun 3, respectively).

---

[8]We would have chosen the 25-by-25 FSS for maximum contrast with 1-by-1 FSS, but models trained with 17-by-17 FSS perform substantially better than those trained with 25-by-25 FSS (*e.g.*, Figure 12).

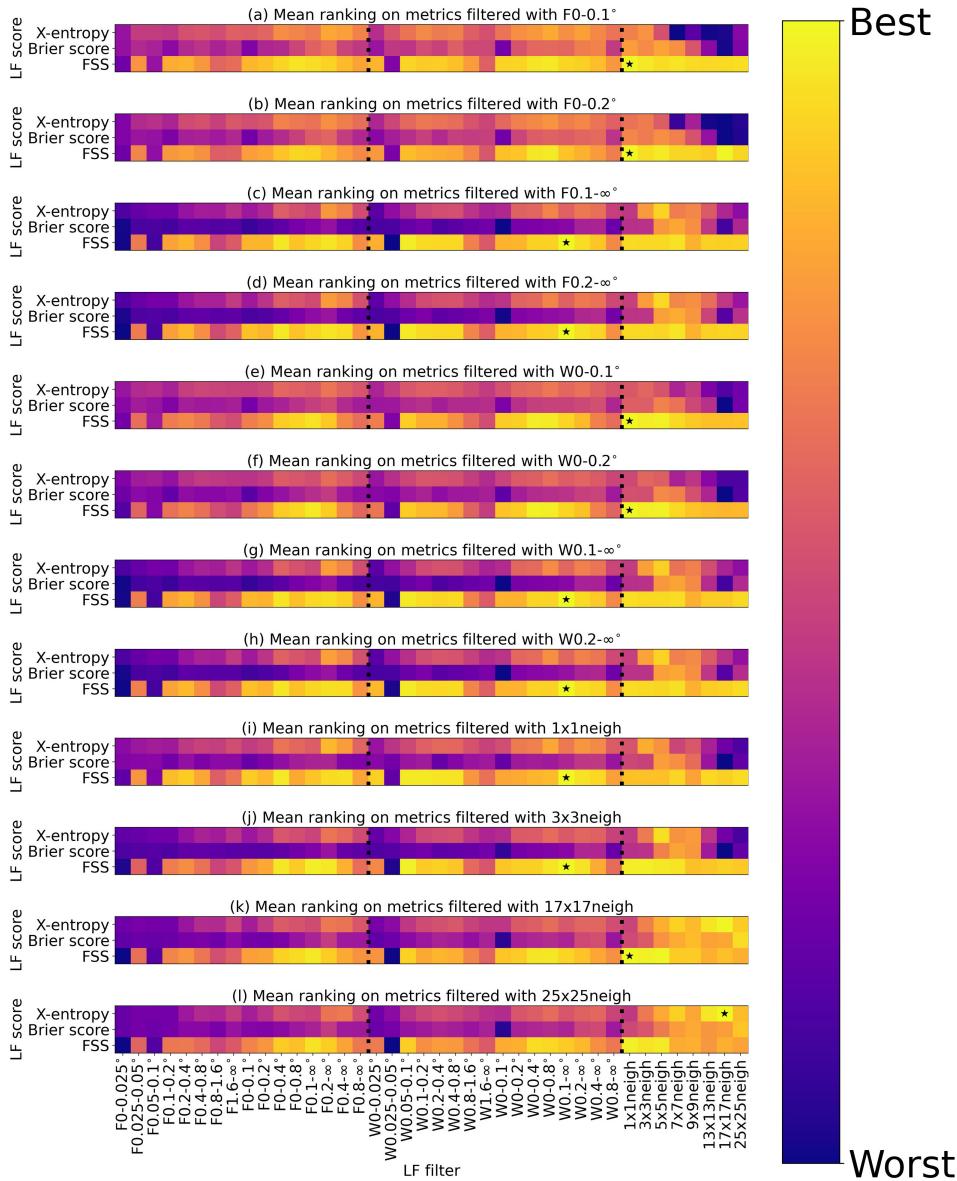

Figure 12: Intermediate analysis. Each panel shows the summary score for 120 different models (those trained with FSS, Brier score, or cross-entropy in the loss function) for verification metrics with one spatial filter, specified in the panel title. "LF score" on the $y$-axis is the score used in the model's loss function, while "LF filter" on the $x$-axis is the spatial filter used in the model's loss function – not to be confused with the filter used in the verification metric. In each panel, there are two dashed vertical lines, separating three groups of models: those trained with a Fourier-based filter (left), those trained with a wavelet-based filter (middle), and those trained with a neighbourhood filter (right). In each panel, the star denotes the best model on the given set of metrics. [a] Summary score on metrics using Fourier decomposition to highlight wavelengths of 0-0.1°. [b-l] Likewise but for metrics with different spatial filters. Only 12 sets of metrics (*i.e.*, metrics involving only 12 of the 40 spatial filters) are shown, for the sake of brevity.

We preface the rest of this section with two notes on the way we interpret Figures 13-14. First, we often use the terms "high risk threshold" and "low risk threshold". A user with a high risk threshold would prefer to be warned only of strong convection, while a user with a low risk threshold would prefer to be warned of any and all convection. Second, we consider forecast probabilities $\geq 0.05$ to be "non-negligible". Although 0.05 is quite a low probability – and would be considered negligible in many applications – (a) our models rarely forecast higher probabilities[9], and (b) in the case studies shown the 0.05-probability contour corresponds well with the edge of the convective area.

---

[9]For the six final models, the histogram of forecast probabilities is shown in Figures 15a-f. Such a skewed distribution is common for rare-event problems.

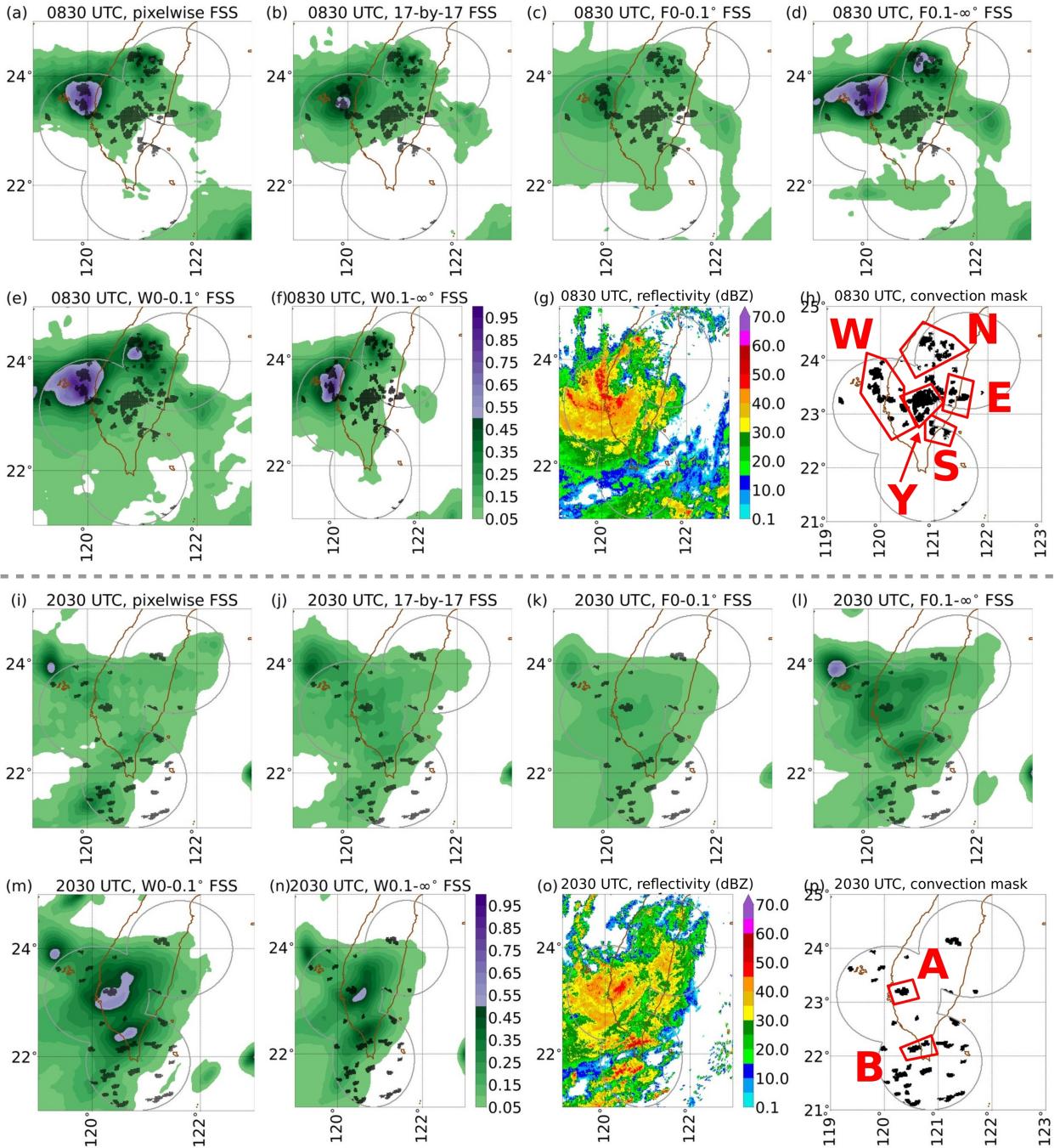

Figure 13: Predictions made by final U-nets (each trained with a loss function involving a different spatial filter on FSS) for Tropical Depression Luis. Formatting (black dots and panel titles) is explained in the captions of Figures 10 and 11. [a-f] Forecast convection probabilities valid at 0830 UTC 23 Aug 2018. [g-h] Composite reflectivity and convection mask valid at 0830 UTC 23 Aug 2018. [i-n] Forecast convection probabilities valid at 2030 UTC 23 Aug 2018. [o-p] Composite reflectivity and convection mask valid at 2030 UTC 23 Aug 2018.

Figure 13 shows two time steps on Aug 23 2018, during the passage of Tropical Depression (TD) Luis. The first time step (0830 UTC; panels a-h) includes areas of strong convection (labeled "W" and "N") and weak convection (labeled "E," "S," and "Y" for eye). The model trained with W0-0.1° FSS (panel e) best captures the structure of the convective area, including both strong and weak convection, but it also produces the most false alarms – *i.e.*, area with probability ≥ 0.05 but no actual convection. Compared to the pixelwise model (panel a), the 17-by-17 model (panel b) has more extremely low probabilities and fewer extremely high probabilities. This is because the 17-by-17 filter smooths the target field (Figure 3h), so the 17-by-17 model is trained with fewer extremely high target values. Otherwise, outside of area W, there are very slight differences between the pixelwise and 17-by-17 models. The F0.1-∞° model (panel d) is similar to the pixelwise model but better at capturing convection in areas E and S. The W0-0.1° model (panel e) is similar to the pixelwise model but better at capturing weak convection in general (areas E, S, and Y). The models not mentioned yet – W0.1-∞° (panel f) and F0-0.1° (panel c) – highlight the strong convection and produce small probabilities (< 0.1) almost everywhere else. Between these two models, the W0.1-∞° model produces much higher probabilities for the strong convection. Overall, for the first time step (0830 UTC) we judge that the W0-0.1° model (panel e) would be best for a low risk threshold, while the W0.1-∞° model (panel f) would be best for a high risk threshold.

At the second time step during TD Luis (2030 UTC; panels i-p), convection is weaker and more spread out. Two areas of strong convection are labeled A and B. Here, the main differences among models are: (a) the F0.1-∞° and W0-0.1° models (panels l, m) are most aggressive, capturing the most convection but also producing the most false alarms; (b) the W0-0.1° model (panel m) captures slightly more convection than the F0.1-∞° model (panel l) and produces substantially higher probabilities for strong convection in areas A and B; (c) of the other four models, the W0.1-∞° model (panel n) produces the highest probabilities for strong convection in areas A and B and has the fewest false alarms. Hence, for the second time step (2030 UTC) we judge that the W0-0.1° model (panel m) would be best for a low risk threshold, while the W0.1-∞° model (panel n) would be best for a high risk threshold.

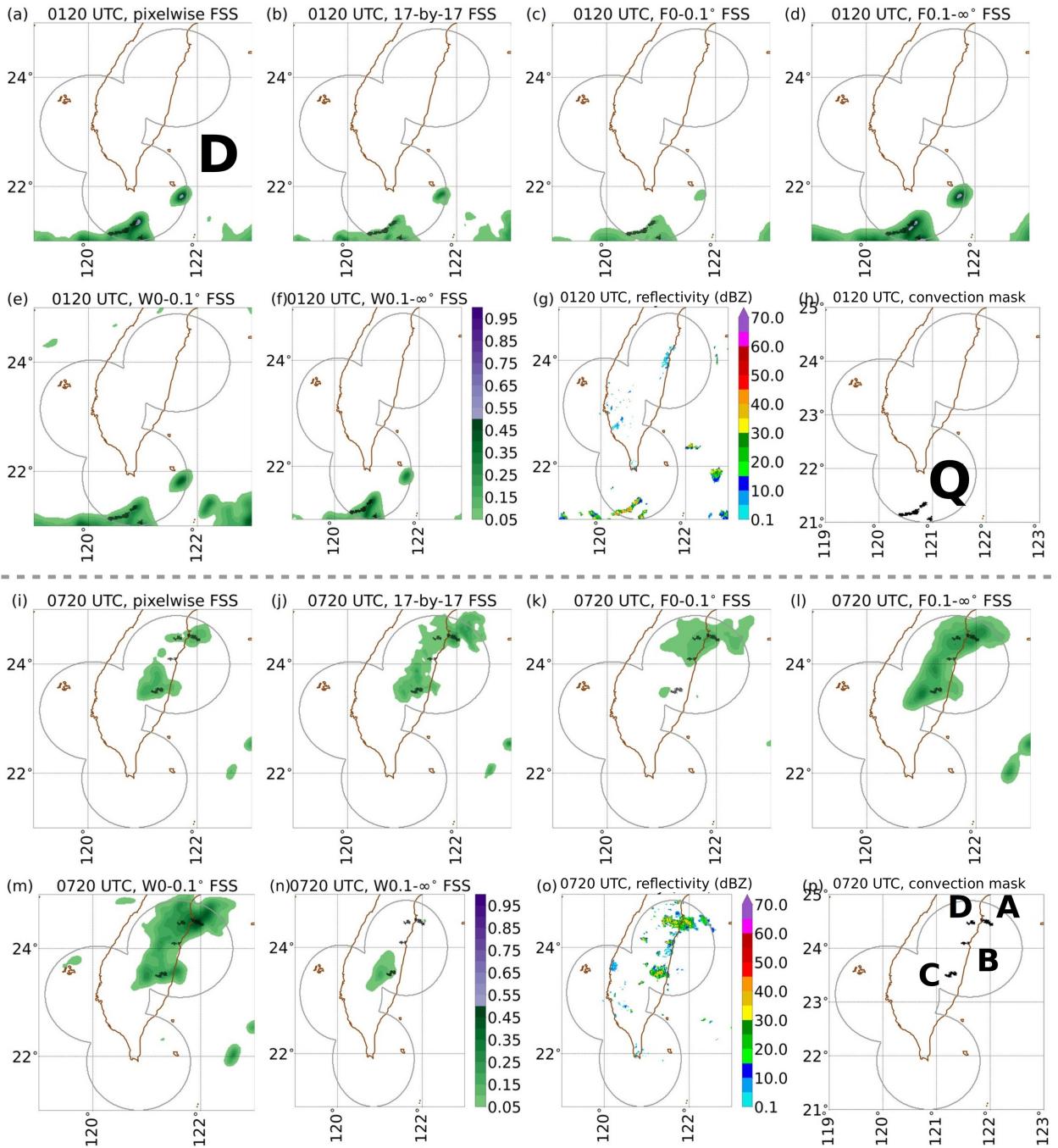

Figure 14: Predictions made by final U-nets (each trained with a loss function involving a different spatial filter on FSS) for summer case. Formatting (black dots and panel titles) is explained in the captions of Figures 10 and 11. [a-f] Forecast convection probabilities valid at 0120 UTC 3 Jun 2018. [g-h] Composite reflectivity and convection mask valid at 0120 UTC 3 Jun 2018. [i-n] Forecast convection probabilities valid at 0720 UTC 3 Jun 2018. [o-p] Composite reflectivity and convection mask valid at 0720 UTC 3 Jun 2018.

Figure 14 shows a summer case (Jun 3 2018) featuring discrete storms, a QLCS at 0120 UTC, and multi-cell clusters at 0720 UTC. At 0120 UTC (panels a-h), where most of the convection is part of the QLCS (labeled "Q"), the pixelwise and F0.1-∞° models (panels a, d) produce the highest probabilities, while the 17-by-17 and F0-0.1° models (panels b, c) produce the lowest probabilities. Also, Fourier-based scale separation produces much more difference in probability magnitudes between the two scales (wavelengths of 0-0.1° and 0.1-∞°; panels c-d) than does wavelet-based scale separation (panels e-f). We believe that this is because, while Fourier and wavelet decomposition generally attribute the same amount of signal to the 0.1-∞° scale (*e.g.*, Figures 4e and 4m), wavelet decomposition generally attributes more signal to the 0-0.1° scale (*e.g.*, Figures 4a and 4i). All models produce a false alarm (labeled "D") associated with convective decay, *i.e.*, the storm existed at forecast-issue time but has since died. At 0720 UTC (panels i-p), there are three multi-cell clusters (labeled "A" through "C") and a discrete storm ("D"). Clusters B and C are convective-initiation events, which only the 17-by-17, F0.1-∞°, and W0-0.1° models capture (panels j, l, m). In general, the W0.1-∞° model (panel n) produces very low probabilities, likely because this model emphasizes larger scales and 0720 UTC features convection at smaller scales. For TD Luis, which featured convection at larger scales, the W0.1-∞° model produced much higher probabilities (panels f, n). Only two models – F0.1-∞° and W0-0.1° (panels l, m) – capture all convective pixels. Overall, for this day (Jun 3) we judge that the F0.1-∞° or W0-0.1° model (panels d-e, l-m) would be best for a low risk threshold. For a high risk threshold, it is difficult to judge which model would be best, since (a) all convection is quite weak, with reflectivity < 45 dBZ; (b) of the remaining models, no model captures all four convective areas at 0720 UTC and all models capture different convective areas.

Supplemental Section 2 discusses a winter case, highlighting two important characteristics of all models. First, the models are poor at capturing weak isolated storms, especially for convective initiation. Second, the models have impressive sharpness for low probabilities – *i.e.*, at times with little to no convection, the models successfully produce very low probabilities throughout the entire domain. In other words, the models do not always produce swaths of elevated probabilities as in the summer case studies (Figures 13-14).

Lastly, we have plotted two verification graphics: the attributes diagram (Hsu and Murphy 1986), which handles probabilistic forecasts, and the performance diagram (Roebber 2009), which handles deterministic forecasts. We summarize the quality of the performance diagram with the area under the curve (AUPD), which ranges from $[0, 1]$ and is positively oriented. We summarize the quality of the attributes diagram with two numbers: reliability (REL), which ranges from $[0, 1]$ and is negatively oriented, and Brier skill score (BSS), which ranges from $(-\infty, 1]$ and is positively oriented. See the original papers for more details. We have added consistency bars to the attributes diagram, following Bröcker and Smith (2007), using 100 bootstrapping iterations[10] and a 95% confidence level. Due to the discretization introduced by binning (*i.e.*, points in the reliability curve are averaged over all cases with a probability of 0.00-0.05, 0.05-0.10, etc.), the reliability curve of a perfect model may deviate slightly from the 1-to-1 line. Consistency bars show how much deviation might be expected.

---

[10]We tried 1000 bootstrapping iterations, but this would have taken too much computing time (> 3 days).

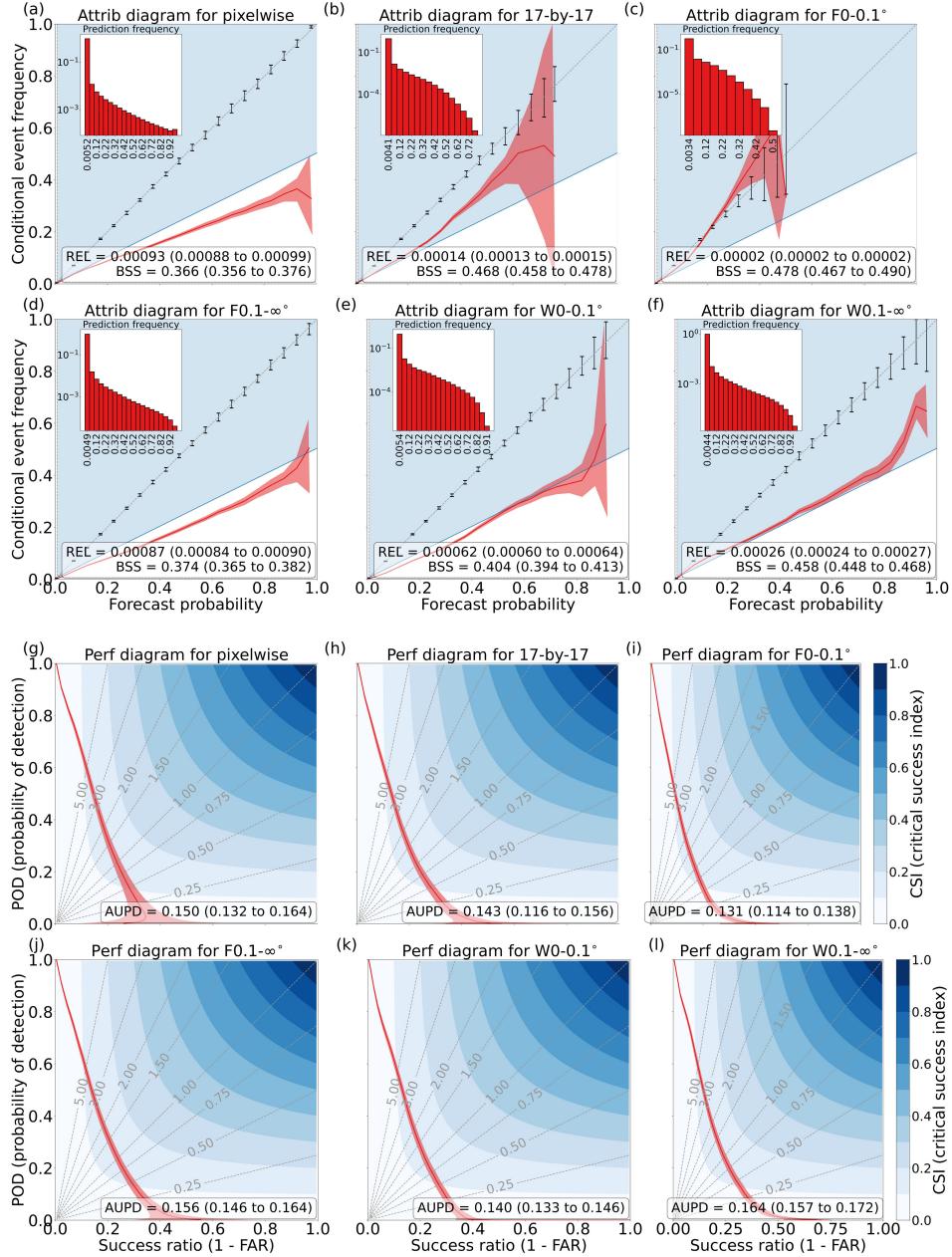

Figure 15: Attributes diagram and performance diagram for each final model. In each panel, the solid red line shows the mean, while the shaded red envelope shows the 95% confidence interval, based on bootstrapping 1000 times. In each attributes diagram, the reliability curve (red) plots conditional event frequency vs. forecast probability. The diagonal grey line is the perfect-reliability line; error bars crossing the perfect-reliability line are consistency bars; the vertical grey line is the climatology line; the horizontal grey line is the no-resolution line; and the blue shading is the positive-skill area, where BSS > 0. REL and BSS are reported as the mean followed by 95% confidence interval. In each performance diagram, the red curve plots POD vs. SR (both defined in Table 1). Each point corresponds to one probability threshold, used to convert probabilistic forecasts to deterministic. Dashed grey lines show the frequency bias ($\frac{POD}{SR}$), for which the optimal value is 1.0, and the blue colour fill shows the CSI, defined in Table 1. AUPD is reported as the mean followed by 95% confidence interval. [a] Attributes diagram for model trained with pixelwise FSS; [b] 17-by-17 FSS; [c] F0-0.1° FSS; [d] F0.1-∞° FSS; [e] W0-0.1° FSS; [f] W0.1-∞° FSS. [g-l] Same as a-f but for performance diagrams.

Figure 15 shows the attributes diagram and performance diagram for each final model – using pixelwise verification, which is the default for these diagrams – based on validation data only. In the attributes diagram (panels a-f), note that all models achieve a positive BSS, even those for which most of the reliability curve lies outside the positive-skill (*i.e.*, positive-BSS) area. This is because most examples fall in the lowest probability bin (as shown by the inset histogram), for which the corresponding (leftmost) point in the reliability curve is inside the positive-skill area. In other words, the lowest probabilities, which occur most often, are well calibrated even if the higher probabilities are not. Note that for rare events it is difficult to calibrate the higher probabilities (*e.g.*, Figure 5 of Gagne et al. 2015, Figure 9 of Gagne et al. 2017, Figure 10f of Lagerquist et al. 2017, Figure 6 of Burke et al. 2020). Additionally, Figure 15 shows three interesting patterns. First, all models with a SELF (panels b-f) have a better attributes diagram than the model with a pixelwise loss function (panel a), in terms of both REL and BSS. For all models except F0.1-∞°, the REL difference and BSS difference versus the pixelwise model is significant at the 95% level, according to a two-sided paired bootstrapping test with 1000 iterations. Second, all models with a SELF (panels h-l) have a nearly equivalent performance diagram to the pixelwise model (panel g), in terms of AUPD (none of the five differences are significant at the 95% level). Third, in our judgement, the W0.1-∞° model has the best combination of probabilistic forecasts (based on the attributes diagram) and deterministic forecasts (based on the performance diagram), with the best performance diagram of all six models and a reliability curve falling completely inside the positive-skill area. The W0.1-∞° model seems preferable to all other models, including the pixelwise model. This is true despite the use of pixelwise verification in Figure 15, which gives an edge to the pixelwise model. Thus, even in a traditional (pixelwise) verification setting, using SELFs can have benefit.

## 7. Discussion and future work

In this section we seek to answer the question from the title: can we integrate SV methods into NN loss functions for atmospheric science? We summarize our insights into this question below, while also considering the question of usefulness.

*Can we integrate neighbourhood methods into loss functions? Is it useful?*

The answer to both questions is a clear yes. Neighbourhood-based verification methods are easy to integrate into loss functions, and it is useful to train NNs to optimize the metrics that users care about, such as the well established FSS. Note that we use a probabilistic version of the FSS – and all other verification scores – instead of thresholding to create binary forecasts. Although we omit thresholding mainly for technical reasons, we believe that thresholding, by destroying the information contained in continuous probabilistic forecasts, would degrade NN performance. Additionally, we learned the following lessons:

- Our sample application (nowcasting convection) does not *resoundingly* demonstrate the usefulness of neighbourhood loss functions, as in many settings the best neighbourhood loss function is a pixelwise one, using the trivial 1-by-1 neighbourhood. We speculate that this is because the double penalty is not a major problem for our application, *i.e.*, 1-hour-ahead forecasting of convection. However, this conclusion is moderated by three other observations. First, Supplemental Section 3 documents a similar experiment but for 2-hour-ahead forecasting, where models trained with the 5-by-5, 9-by-9, and 13-by-13 neighbourhood filters perform extremely well and much better than any model trained with a pixelwise loss function[11] (*c.f.* Figure 12 and Supplemental Figure S4). Second, models trained with non-trivial neighbourhood loss functions, such as the 17-by-17 FSS, produce much better-calibrated probabilities. Third, for applications where the double penalty is a major problem, such as detecting the locations of fronts (Justin et al. 2022) and wildfires (Earnest et al. 2022), neighbourhood loss functions still appear to be crucial.
- We explored a much wider range of neighbourhood loss functions than previous work (FSS only). We also found – at least for our application – that neighbourhood loss functions including the FSS are better than those including other verification metrics. There is good reason to believe that this conclusion holds for predicting rare events in general.
- One should be careful when using large neighbourhood sizes, as this may trigger the excessive-freedom problem, manifesting as checkerboard artifacts in the predictions.

---

[11] All three of these neighbourhood-based models perform much better than that trained with the pixelwise FSS. However, this comparison is not very stringent, because the model trained with the pixelwise FSS failed to converge. As discussed in Supplemental Section 3, in the interest of fairness and computing time, we do not retrain models that failed to converge.

*Can we integrate scale separation into loss functions? Is it useful?*

The answer here is more nuanced. Implementing a spectral filter in the loss function is difficult, and we needed to modify our approach to yield useful results, putting the spectral filter outside the loss function. However, the final approach still integrates scale separation into NN-training and shows different benefits than neighbourhood loss functions. We learned the following lessons:

- It is possible to implement spectral filters with Fourier decomposition directly in the loss function. However, this leads to the excessive-freedom problem, as the NN can "go haywire" at scales removed by the spectral filter.
- It is technically challenging to implement spectral filters with wavelet decomposition directly in the loss function, and even if the technical (coding) challenges were overcome, this would lead to the excessive-freedom problem as well.
- The scale-separation approach has benefits not achieved by the neighbourhood approach. First, the W0.1-$\infty$° model (trained to optimize FSS at wavelengths $\geq 0.1$°) produces better probabilistic forecasts and comparable deterministic forecasts to the pixelwise model, even when assessed with pixelwise verification. Second – considering results from objective verification and the subjective case studies – the best models appear to be pixelwise and W0-0.1° for a low risk threshold, W0.1-$\infty$° for a high risk threshold. Of these three models, two are trained with scale separation.

In addition to the technical details and experimental results summarized above, we provide Python code to implement SELFs. We believe that SELFs should be part of the developer's toolbox for training NNs – and other methods with loss functions, including other machine-learning models and data assimilation – to optimize the performance measures that users truly care about. However, we have only scratched the surface in discovering what these tools can do and how best to use them. Future work will proceed along three lines. First, we will develop temporally enhanced loss functions for predicting time series, which would ideally make predictions more temporally smooth (not varying erratically between time steps) and prevent the temporal version of the double penalty. Second, we will combine physical constraints (already used in loss functions for some geoscience applications) with SELFs. Third, we will investigate SELFs for more applications to build an understanding of which methods are best for which applications.

*Acknowledgments.* We thank the Taiwan CWB for providing the satellite and radar data used herein. We also thank an anonymous reviewer for very helpful comments, including the suggestion of additional experimentation. This work was partially supported by the NOAA Global Systems Laboratory, Cooperative Institute for Research in the Atmosphere, and NOAA Award Number NA19OAR4320073. Author Ebert-Uphoff's work was partially supported by NSF AI Institute grant ICER-2019758 and NSF grant OAC-1934668.

*Data availability statement.* Input data (satellite and radar images) are available upon request from the authors, as well as trained versions of all U-net models. We used version 2.0.0 of ML4convection (doi:10.5281/zenodo.6359915) – a Python library managed by author Lagerquist – to train and verify all models in this work. More importantly, we have created a Colab notebook (`https://colab.research.google.com/drive/1g8xrjOv0Vs4ryBvPxU7bqHIXoluI_Ses?usp=sharing`) with all the relevant code (independent of ML4convection), providing an end-to-end resource that people can use to experiment with our SELFs and implement their own.

Sobash, R., J. Kain, D. Bright, A. Dean, M. Coniglio, and S. Weiss, 2011: Probabilistic forecast guidance for severe thunderstorms based on the identification of extreme phenomena in convection-allowing model forecasts. *Weather and Forecasting*, **26 (5)**, 714–728, URL https://doi.org/10.1175/WAF-D-10-05046.1.

Starzec, M., C. Hometer, and G. Mullendore, 2017: Storm Labeling in Three Dimensions (SL3D): A volumetric radar echo and dual-polarization updraft classification algorithm. *Monthly Weather Review*, **145 (3)**, 1127–1145.

Stengel, K., A. Glaws, D. Hettinger, and R. King, 2020: Adversarial super-resolution of climatological wind and solar data. *Proceedings of the National Academy of Sciences*, **117 (29)**, 16 805–16 815, URL https://doi.org/10.1073/pnas.1918964117.

Versaci, F., 2021: WaveTF: A fast 2D wavelet transform for machine learning in Keras. *Proceedings of the International Conference on Pattern Recognition*, 605–618, URL https://doi.org/10.1007/978-3-030-68763-2_46.

Wang, C., C. Xu, C. Wang, and D. Tao, 2019: Perceptual adversarial networks for image-to-image transformation. *Transactions on Image Processing*, **27 (8)**, 4066–4079, URL https://doi.org/10.1109/TIP.2018.2836316.

Wang, L., Y. Zhang, and J. Feng, 2005: On the Euclidean distance of images. *IEEE Transactions on Pattern Analysis and Machine Intelligence*, **27 (8)**, 1334–1339, URL https://doi.org/10.1109/TPAMI.2005.165.

Wang, Z., and A. Bovik, 2009: Mean squared error: Love it or leave it? A new look at signal fidelity measures. *IEEE Signal Processing Magazine*, **26 (1)**, 98–117, URL https://doi.org/10.1109/MSP.2008.930649.

Wang, Z., A. Bovik, H. Sheikh, and E. Simoncelli, 2004: Image quality assessment: from error visibility to structural similarity. *IEEE Transactions on Image Processing*, **13 (4)**, 600–612, URL https://doi.org/10.1109/TIP.2003.819861.

Weusthoff, T., F. Ament, M. Arpagaus, and M. Rotach, 2010: Assessing the benefits of convection-permitting models by neighborhood verification: Examples from MAP D-PHASE. *Monthly Weather Review*, **138 (9)**, 3418–3433, URL https://doi.org/10.1175/2010MWR3380.1.
50

# Supplemental material

## 1. Procedures for implementing spectral filters

In this section we provide a detailed description of the procedures for Fourier and wavelet decomposition, so that others may reproduce our results and use these methods in their own research. We also provide code (see link in "Data-availability Statement").

Let $\delta_{\min}$ and $\delta_{\max}$ be the minimum and maximum spatial resolutions allowed through the spectral filter. Thus, $\lambda_{\min} = 2\delta_{\min}$ and $\lambda_{\max} = 2\delta_{\max}$ are the minimum and maximum wavelengths, while $\nu_{\min} = \frac{1}{\lambda_{\max}}$ and $\nu_{\max} = \frac{1}{\lambda_{\min}}$ are the minimum and maximum wavenumbers.

*a. Fourier decomposition*

For one target field – *i.e.*, the convection mask at one time step – the procedure is described below. See Figure 7 in the main text for a schematic illustration, where the spectral filter allows wavelengths from $\lambda_{\min} = 0.5°$ to $\lambda_{\max} = 2°$.

1. **Tapering in spatial domain.** The Fourier transform assumes a periodic domain, so if we analyze a finite spatial domain, the Fourier transform implicitly assumes that the domain is toroidal. In other words, the Fourier transform assumes that the top and bottom rows are adjacent, also that the left and right columns are adjacent – causing spurious patterns in the wavelength domain. To prevent these spurious patterns, we taper the data before the forward transform, then undo tapering after the inverse transform (Step 6). Specifically, we triple the grid size from $205 \times 205$ to $615 \times 615$ and fill the new pixels with zeros (indicating no convection).

2. **Windowing in spatial domain.** Even after tapering, the Fourier transform's assumption of a periodic domain is problematic. As shown in Figure 8.7 of Stull (1988), a smoothly varying pattern in a finite spatial domain can appear as a sawtooth pattern in the wavelength domain, containing many spurious wavelengths called "red noise". In general, if data in a finite spatial domain are not processed with an appropriate window, the result is leakage in the wavelength domain, where the estimated amount of signal (*i.e.*, amplitude of Fourier coefficient) at one wavelength is contaminated by neighbouring wavelengths. See Section 8.4.3 of Stull (1988) for details.

We use a 2-D, radially symmetric version of the Blackman-Harris window (Harris 1978), defined as:

$$\begin{cases} w(r_g) = 0.42 - 0.5 \cos(\pi \left[1 + \frac{r_g}{R}\right]) + 0.08 \cos(2\pi \left[1 + \frac{r_g}{R}\right]), & \text{if } r_g \leq R; \\ w(r_g) = 0, & \text{if } r_g > R. \end{cases} \quad (1)$$

$r_g$ is the distance of grid point $g$ from the domain center; $R$ is the maximum distance; and $w$ is the resulting weight. We set $R$ to 307 grid points, the half-width of the 615-by-615 tapered grid. We multiply the weight matrix elementwise with the original field (predictions or observations).

3. **Forward transform.** Use the Fourier transform to convert the spatial field into the wavelength domain. For a spatial domain of $N \times N$ grid points, the Fourier transform outputs an $N \times N$ complex-valued coefficient matrix in the wavelength domain. For each spatial direction this matrix contains one coefficient for the infinite wavelength (zero frequency), $\frac{1}{2}(N-1)$ coefficients for positive wavelengths, and $\frac{1}{2}(N-1)$ coefficients for negative wavelengths. We split the coefficient matrix into two: one containing the magnitudes, the other containing the phases, of the complex-valued coefficients.

4. **Filtering in wavelength domain.** The naïve approach would be to zero out coefficients at undesired wavelengths (*i.e.*, use a rectangular filter), but this causes artifacts in the spatial domain after the inverse transform. Instead, we use a second-order Butterworth (1930) filter, which decreases the magnitude of all coefficients (the phase remains unchanged) but especially those at undesired wavelengths. We use a band-pass filter, which is a superposition of low-pass and high-pass filters. The low-pass filter admits low frequencies (large wavelengths), while the high-pass filter admits high frequencies (small wavelengths). For each wavenumber pair $(k, l)$ – where $k$ is the zonal wavenumber and $l$ is the meridional wavenumber, both in units of m$^{-1}$ – the gain of the Butterworth filter is defined as[1]:

$$\begin{cases} g_{\text{low}}(k,l) = \dfrac{1}{1+\left(\dfrac{\sqrt{k^2+l^2}}{\nu_{\max}}\right)^{2\alpha}}; \\ g_{\text{high}}(k,l) = 1 - \dfrac{1}{1+\left(\dfrac{\sqrt{k^2+l^2}}{\nu_{\min}}\right)^{2\alpha}}. \end{cases} \qquad (2)$$

$\sqrt{k^2+l^2}$ is the total wavenumber; $\alpha$ = 2 is the order of the filter; $g_{\text{low}} \in [0,1]$ is the gain for the low-pass filter; and $g_{\text{high}} \in [0,1]$ is the gain for the high-pass filter. For each wavelength pair we multiply the magnitude of the Fourier coefficient by $g_{\text{low}}$ to implement the low-pass filter, then $g_{\text{high}}$ to implement the high-pass filter.

5. **Inverse transform.** Use the inverse Fourier transform to convert the coefficients back to the spatial domain. If the coefficients were not filtered, the inverse transform would exactly recover the original spatial field. With filtered coefficients, the inverse transform recovers the part of the spatial field corresponding to the desired wavelengths.

6. **Undo tapering.** Remove zero-padding from the reconstructed spatial field.

---

[1] https://www.originlab.com/doc/Origin-Help/2DFFT-Filter-Algorithm#Butterworth

*b. Wavelet decomposition*

For one target field – *i.e.*, the convection mask at one time step – the procedure is described below. See Figure 8 in the main text for a schematic illustration, where the spectral filter allows wavelengths from $\lambda_{\min} = 0.1°$ to $\lambda_{\max} = 0.4°$. For more information on wavelet decomposition, see Kumar and Foufoula-Georgiou (1994), Vidakovic and Müller (1994), and Versaci (2021). For wavelet decomposition involving 2-D images, we especially recommend the illustrations in Versaci (2021), although this paper serves only partly as a theoretical treatment of wavelet decomposition, devoting much space to technical documentation for the WaveTF software library.

1. **Tapering.** Taper the field by zero-padding. The wavelet transform needs both grid dimensions to be an integer power of 2, so we taper the grid from $205 \times 205$ to $256 \times 256$. As in Fourier decomposition, we fill the new pixels with zeros (indicating no convection).

2. **Forward transform.** Use the wavelet transform to convert the tapered field from the spatial domain to the wavelength domain. We use the WaveTF library (Versaci 2021) in Python, which offers the choice between two wavelet types: Haar and Daubechies. We use the Haar wavelet because it is less computationally expensive. For a spatial domain of $2^K \times 2^K$ grid points, the wavelet transform outputs coefficients at $K$ levels. At each level there are four sets of coefficients: LL (representing the low frequency, or mean, in both the vertical and horizontal), LH (representing the mean in the horizontal and high frequency [details] in the vertical), HL (the opposite), and HH (representing details in both the vertical and horizontal). At each successive level, the grid size halves while the wavelengths represented double.

3. **Filter out coefficients at undesired wavelengths.**

(a) Zero out LL coefficients for all wavelengths $> \lambda_{\max}$. In Figure 8 of the main text, this is done for levels 5 through 8 (where LL coefficients are for wavelengths of 0.8°, 1.6°, 3.2°, and 6.4°).

(b) At each level $k$ with wavelengths in the allowed range, starting with the deepest level, use the inverse wavelet transform to reconstruct LL coefficients at level $k$ from all coefficients at level $k+1$. In Figure 8 of the main text, this is done for levels 4, then 3.

(c) At each level $k$ with wavelengths $< \lambda_{\min}$, starting with the deepest level, use the inverse wavelet transform to reconstruct LL coefficients at level $k$ from all coefficients at level $k+1$, then zero out all detail coefficients (LH, HL, and HH) at level $k$. In Figure 8 of the main text, this is done for levels 2, then 1.

4. **Inverse transform.** Use the inverse wavelet transform to convert the filtered level-1 coefficients back to the spatial domain. The inverse wavelet transform, like the inverse Fourier transform, would exactly recover the original spatial field if coefficients were not filtered. With filtered coefficients, the inverse transform recovers the part of the spatial field corresponding to the desired wavelengths.

5. **Undo tapering.** Remove zero-padding from the reconstructed spatial field.

## 2. Additional case study

Figure S1 shows a winter case, with two time steps on Jan 25 2018: 1230 and 2230 UTC. This case is almost trivial, as both time steps feature only one discrete storm – both examples of convective initiation, where the storm exists at the valid time ($t_0 + 1$ hour) but not at the forecast-issue time ($t_0$) – and no probabilities $\geq 0.05$ from any model. However, this case illustrates two important characteristics of all models, including the pixelwise model and those with SELFs. First, the models are poor at capturing weak isolated storms, especially for convective initiation. Second, the models have excellent sharpness for low probabilities – *i.e.*, when there is little to no convection in the domain, they are capable of predicting negligible probabilities ($< 0.05$) everywhere.

## 3. Additional experiment (2-hour lead time)

We conduct an experiment similar to that shown in the main body, but with a lead time of 2 hours instead of 1 hour. We hypothesized that SELFs would have a greater benefit at longer lead times, because longer lead times involve more advection (storm motion), which causes a greater spatial offset between predicted and observed convection, thus making the double penalty a bigger problem.

We train U-nets with 120 of the 336 loss functions used in the 1-hour experiment: those involving the FSS, Brier score, or cross-entropy. Based on results from the 1-hour experiment (Section 6a in main body), we believe that the other scores (IOU, Dice coefficient, CSI, Peirce score, Heidke score, and Gerrity score) are not worth considering here; including them would needlessly consume computing resources.

As for the 1-hour experiment, we separate our analysis into three steps: the preliminary, intermediate, and final analyses. In the preliminary and final analyses, where a conclusion is supported by a single time step, we have ensured that the conclusion is representative of the entire corresponding day.

### a. Preliminary analysis

In the 1-hour experiment we used the preliminary analysis to dismiss a large swath of loss functions (those involving the IOU, Dice coefficient, CSI, Peirce score, Heidke score, or Gerrity score) as inappropriate for the task of predicting convection, thus paring down the number of models for the intermediate analysis. Here, in the 2-hour experiment, we use the preliminary analysis to gain a qualitative understanding of the properties of each model, before conducting the fully quantitative intermediate analysis.

Figure S2 shows predictions made by models trained with neighbourhood loss functions, valid at 2200 UTC 2 Jun 2017, the same time step used in the preliminary analysis for the 1-hour experiment. Analyzing the models' predictions for this day yielded the following key results:

- Some models (panels i, j, l, o) make very poor predictions, because the models did not converge[2]. Specifically, the four models in question reached their minimum validation loss after (1, 2, 2, 2) epochs respectively, despite being trained for 30 epochs thereafter. In the interest of computing time, we left such non-converging models "as is," instead of training with different random seeds until we obtained a converging model. We tried the latter approach and found that models may need to be trained up to 30 times before convergence. In the interest of fairness, if we trained *some* models 30 times and took the best version, we would need to retrain *all* models 30 times and take the best version. This would involve training $120 \times 30 = 3600$ models, which would take weeks on our computing resources (2000 CPU cores and 800 GPU cores).

- In the 1-hour experiment we did not find non-converging models. We believe that non-convergence is a problem for the 2-hour experiment because the prediction task is more difficult, leading to fewer acceptable local minima.

---

[2]The loss function depends on all NN parameters (weights and biases), and each U-net in this experiment has 542 053 parameters. In this very high-dimensional space, the loss function has many local minima, some of which are better than others. The same NN with different random seeds – *i.e.*, different initializations of the 542 053 parameters, leading to different starting points in the loss-function space – can reach very different local minima. When a model reaches a poor local minimum early in the training procedure and gets stuck there, we say that the model "did not converge".

- The spatially averaged forecast probability generally increases with neighbourhood size, *i.e.*, along each row of Figure S2. We believe that this is because larger neighbourhood sizes more effectively avoid the double penalty, which is a bigger problem at 2-hour lead time than at 1-hour lead time, due to the amount of storm motion that the U-net must implicitly forecast.

- Other than the non-converging models, all models produce reasonable probabilities –*i.e.*, no large areas of near-zero probability that coincide with a lot of convection, no large areas of near-100% probability that coincide with no convection at all.

Figure S3 is analogous to Figure S2 but for models trained with scale-separation loss functions. Key results are listed below:

- The three models with wavy artifacts in the probability field – F0.1-$\infty$° FSS (panel j), W0-0.1° FSS (panel k), and W0.1-$\infty$° FSS (panel l) – are all non-converging. They reached their minimum validation loss after (1, 2, 1) epochs respectively, despite being trained for 30 epochs thereafter.

- Models trained with the Brier score (panels a-h) produce negligible probabilities ($< 0.05$) everywhere; models trained with cross-entropy (panels q-x) produce negligible probabilities almost everywhere; and models trained with the FSS (panels i-p) are the only ones that produce large areas of non-negligible probabilities. These conclusions also apply to scale-separation-based loss functions for the 1-hour experiment (Figure 11 in main body).

*b. Intermediate analysis*

As in the 1-hour experiment, we use the intermediate analysis to select a handful of models for the final analysis, which consists of in-depth case studies. We use the same procedure, yielding one summary score for each of 120 models on each of the 40 spatial filters used in the verification metrics. Results for the 2-hour lead time are summarized below and in Figure S4.

- For 37 of the 40 filters, the best model includes the FSS, rather than the Brier score or cross-entropy, in the loss function. Thus, as for the 1-hour lead time, the FSS appears to be the most appropriate score for producing a skillful model.

- For the other 3 filters, the best model is trained with 5-by-5 cross-entropy.

- For 30 of the 40 filters, the best model is trained with a neighbourhood-based, rather than scale-separation-based, FSS. The specific neighbourhood sizes are $5 \times 5$ (ranks best for 14 verification filters), $9 \times 9$ (best for 8 filters), and $13 \times 13$ (also ranks best for 8 filters).

- For 7 of the 40 filters, the best model is trained with a scale-separation-based FSS. The specific filters are F0-0.8° (ranks best for 5 verification filters), W0.4-∞° (ranks best for 1 filter), and F0.1-0.2° (also ranks best for 1 filter).

For the final analysis, we select the model trained with pixelwise FSS as a baseline, as in the 1-hour experiment, because pixelwise FSS is not spatially enhanced. We also select the following models to compare against the baseline: those trained with 5-by-5 FSS, 9-by-9 FSS, 13-by-13 FSS, F0-0.8° FSS, and 5-by-5 cross-entropy.

*c. Final analysis*

In this section we analyze the properties of forecasts produced by the six selected models – trained with pixelwise FSS, 5-by-5 FSS, 9-by-9 FSS, 13-by-13 FSS, F0-0.8° FSS, and 5-by-5 cross-entropy – on the same case studies as for the 1-hour experiment. However, we do not show the winter case study (analogue to Figure S1), because the results are the same – *i.e.*, all models produce negligible probabilities throughout the domain.

Figure S5 shows two time steps during the passage of Tropical Depression (TD) Luis. As in Figure 13 of the main body, for ease of interpretation, we have labeled areas of strong (W, N) and weak (Y, E, S) convection at the first time step and two areas of strong convection (A, B) at the second time step. At the first time step (0830 UTC; panels a-h), the model trained with 9-by-9 FSS (panel c) best captures the structure of the convective area, while producing substantially higher probabilities for the strong convection than the weak convection. The model trained with 5-by-5 FSS (panel b) produces higher probabilities for the whole convective area, especially the strong convection, but also produces many false alarms, most notably in the southern third of the domain (south of $\sim$22.5 °N). The models trained with 13-by-13 FSS (panel d) and F0-0.8° FSS (panel e) are similar to that trained with 5-by-5 FSS, producing many false alarms in the southern third, but also produce much lower probabilities for the strong convection. Of the two models heretofore not mentioned, that trained with pixelwise FSS produces near-100% probabilities through most of the domain, rendering it useless. Finally, the model trained with 5-by-5 cross-entropy (panel f) highlights the strong convection and is the only model producing negligible probabilities ($<$ 0.05) for the weak convection. Overall, for the first time step (0830 UTC) we judge that the model trained with 5-by-5 cross-entropy would be most appropriate for users with a high risk threshold, while that trained with 9-by-9 FSS would be most appropriate for a low risk threshold.

Results for the second time step during TD Luis (2030 UTC; panels i-p) are qualitatively similar. The model trained with 5-by-5 FSS captures the whole convective area and produces the highest probabilities for both the strong and weak convection (except the pixelwise model, which again produces perversely high probabilities), making it most appropriate for users with a low risk threshold. The model trained with 5-by-5 cross-entropy captures the two highlighted areas of strong convection and ignores the mostly-weak convection southeast of Taiwan, making it most appropriate for a high risk threshold.

Figure S6 shows two time steps on Jun 3 2018, a more typical summer day, featuring convection at various scales smaller than a tropical cyclone. At the first time step (0120 UTC; panels a-h), all models trained with a spatially enhanced FSS (panels b-e) produce negligible probabilities for the QLCS (labeled "Q") and non-negligible probabilities for the convective-decay event (labeled "D"), constituting a false negative and false positive respectively. Meanwhile, the model trained with 5-by-5 cross-entropy produces negligible probabilities everywhere with defined target values (*i.e.*, inside the 100-km range rings). Since all convection in the domain is quite weak, we conclude for this time step that the model trained with 5-by-5 cross-entropy would be best for users with a high risk threshold. For users with a low risk threshold, all models are inappropriate, since that trained with pixelwise FSS produces a large swath of perversely high probabilities and the others produce negligible probabilities throughout the convective area. At the second time step (0720 UTC; panels i-p), the structure of the convective area is captured best by models trained with 5-by-5 FSS (panel j), 9-by-9 FSS (panel k), and F0-0.8° FSS (panel m). The structure is mostly captured by the model trained with 5-by-5 cross-entropy as well, which produces substantially lower probabilities than the other models mentioned. Since all convection in the domain is quite weak, again we conclude that the model trained with 5-by-5 cross-entropy would be most appropriate for users with a high risk threshold. For users with a low risk threshold, we deem that the model trained with 5-by-5 FSS would be most appropriate, since it captures the structure of the convective area as well as any other model but produces higher probabilities.

Lastly, Figure S7 shows the attributes diagram and performance diagram for each final model – using pixelwise verification, which is the default for these diagrams – based on validation data. As discussed in the main body, the attributes (performance) diagram handles probabilistic (deterministic) forecasts. We summarize the quality of the attributes diagram with the positively oriented Brier skill score (BSS) and negatively oriented reliability (REL); we summarize the quality of the performance diagram with the positively oriented area under the curve (AUPD).

In terms of probabilistic forecasts (panels a-f), the best model is that trained with 5-by-5 cross-entropy. Comparing this model to the others, all five BSS differences and all five REL differences are significant at the 95% confidence level, according to a two-sided paired bootstrapping test with 1000 iterations. In the case studies explored above, the model trained with 5-by-5 cross-entropy also appeared to be best for users with a high risk threshold; thus, objective spatial evaluation (Supplemental Section 2b), subjective evaluation (earlier in this section), and objective pixelwise evaluation (Figure S7) highlight this model as one of the most attractive options. In terms of deterministic forecasts (panels g-l), the best model is again that trained with 5-by-5 cross-entropy. However, AUPD differences between this model and the other spatially enhanced models (all except pixelwise FSS) are not significant at the 95% level. Lastly, although the model trained with 5-by-5 FSS appeared in case studies to the best for users with a low risk threshold, it has the worst probability calibration of all models trained with a SELF (panels b-f). The four REL differences and BSS differences are significant at the 95% level. However, it is important to remember that Figure S7 uses pixelwise evaluation. Sometimes pixelwise evaluation and subjective evaluation are complimentary, as for the model trained with 5-by-5 cross-entropy, but sometimes this is not the case. Such disagreements highlight the importance of holistic model verification, including both subjective and objective methods.

# List of Figures

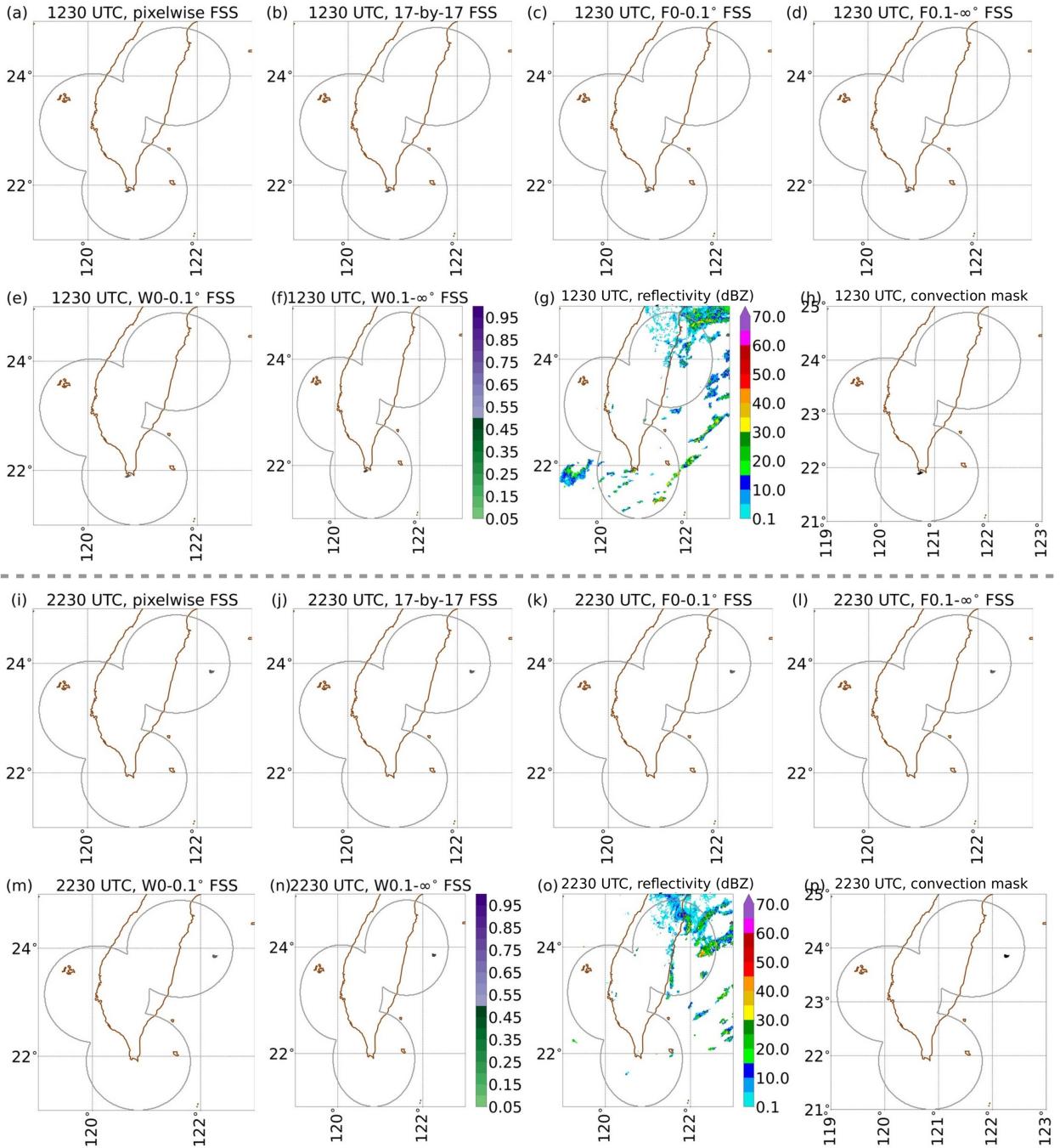

Figure S1: Predictions made by final 1-hour U-nets (each trained with a loss function involving a different spatial filter on FSS) for winter case. Formatting (black dots and panel titles) is explained in the captions of Figures 10-11 in the main body. [a-f] Forecast convection probabilities valid at 1230 UTC 25 Jan 2018. [g-h] Composite reflectivity and convection mask valid at 1230 UTC 25 Jan 2018. [i-n] Forecast convection probabilities valid at 2230 UTC 25 Jan 2018. [o-p] Composite reflectivity and convection mask valid at 2230 UTC 25 Jan 2018.

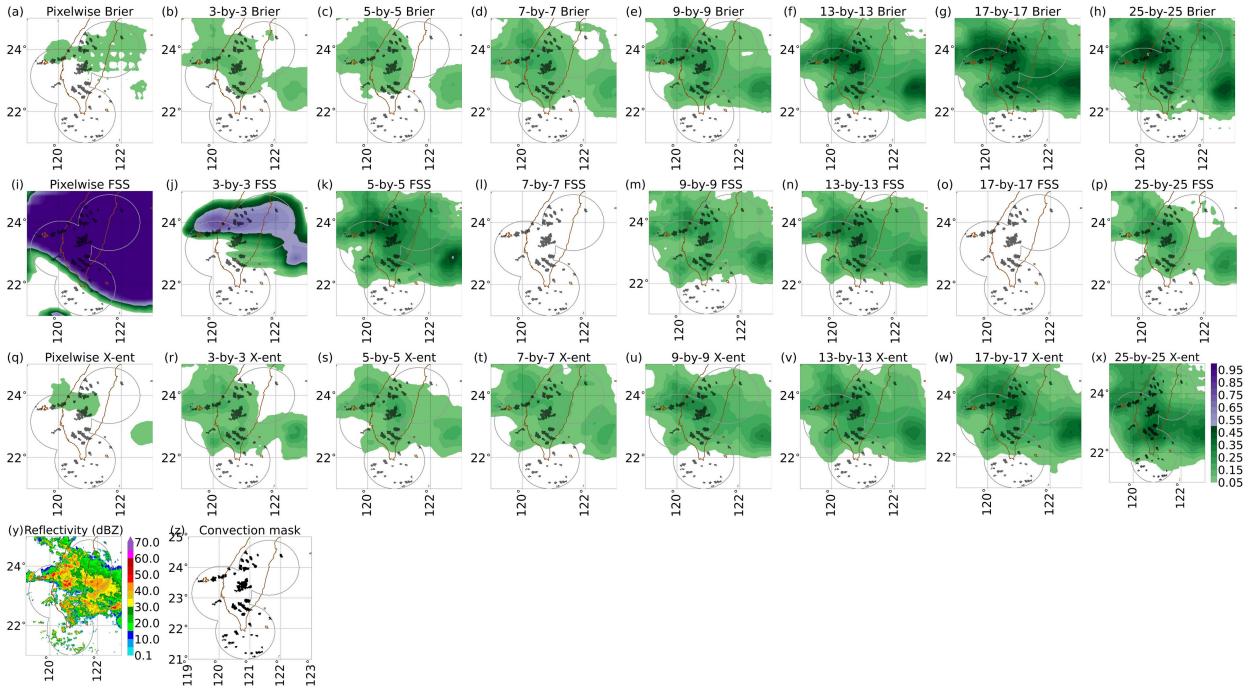

Figure S2: Predictions valid at 2200 UTC 2 Jun 2017, made by U-nets trained with different neighbourhood loss functions. Formatting is explained in the caption of Figure 10 in the main body. [a-h] Convection probabilities forecast by U-nets trained to optimize the Brier score with different neighbourhoods. [i-p] Same but for FSS. [q-x] Same but for cross-entropy. [y] Composite (column-maximum) radar reflectivity valid at 2200 UTC. [z] Convection mask valid at 2200 UTC.

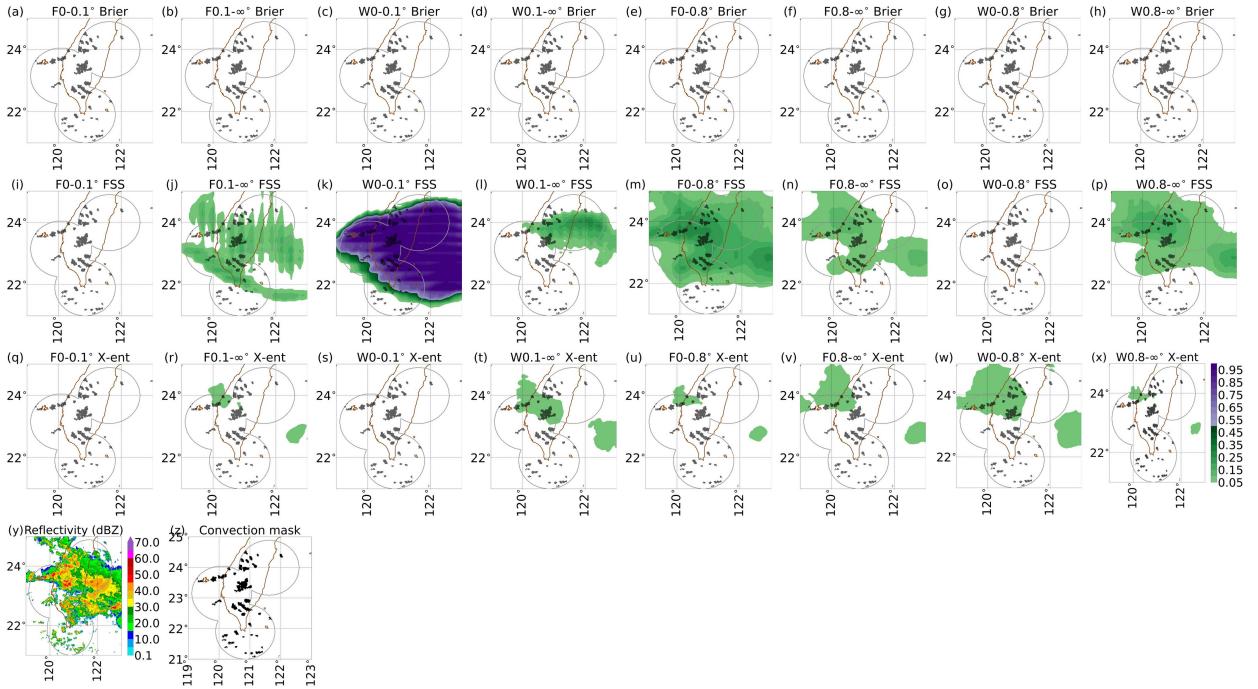

Figure S3: Predictions valid at 2200 UTC 2 Jun 2017, made by U-nets trained with different scale-separation loss functions. Formatting is explained in the caption of Figure 10 in the main body. [a-h] Convection probabilities forecast by U-nets trained to optimize the Brier score with different spectral filters. For example, the filter "F0-0.1°" uses Fourier decomposition to highlight wavelengths $\leq 0.1°$, and "W0.8-$\infty$°" uses wavelet decomposition to highlight wavelengths $\geq 0.8°$. [i-p] Same but for FSS. [q-x] Same but for cross-entropy. [y] Composite (column-maximum) radar reflectivity valid at 2200 UTC. [z] Convection mask valid at 2200 UTC.

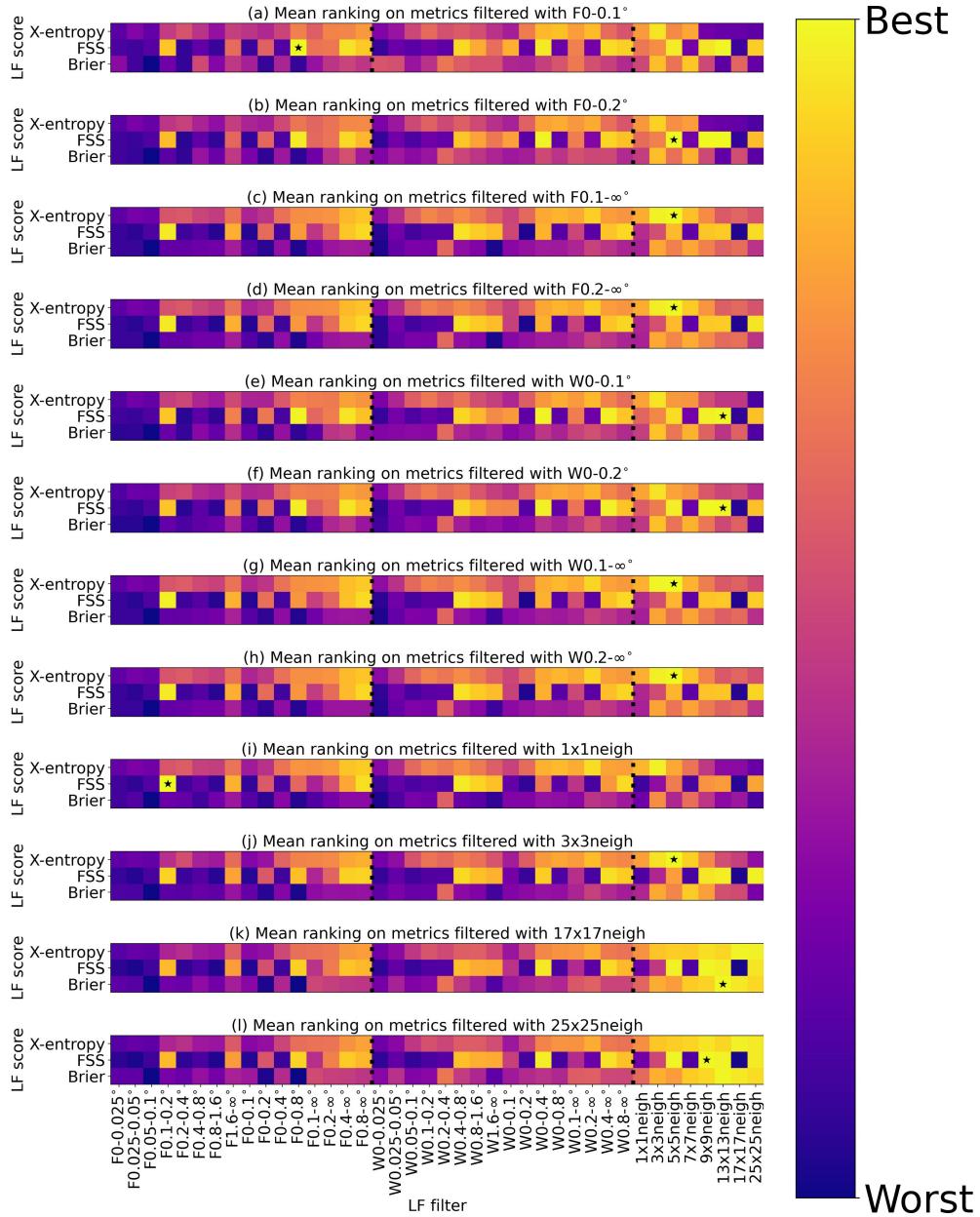

Figure S4: Intermediate analysis. Each panel shows the summary for all 120 models for verification metrics with one spatial filter. Formatting is explained in the caption of Figure 12 in the main body. [a] Summary score on metrics using Fourier decomposition to highlight wavelengths of 0-0.1°. [b-l] Likewise but for metrics with different spatial filters. Only 12 sets of metrics (*i.e.*, metrics involving only 12 of the 40 spatial filters) are shown, for the sake of brevity.

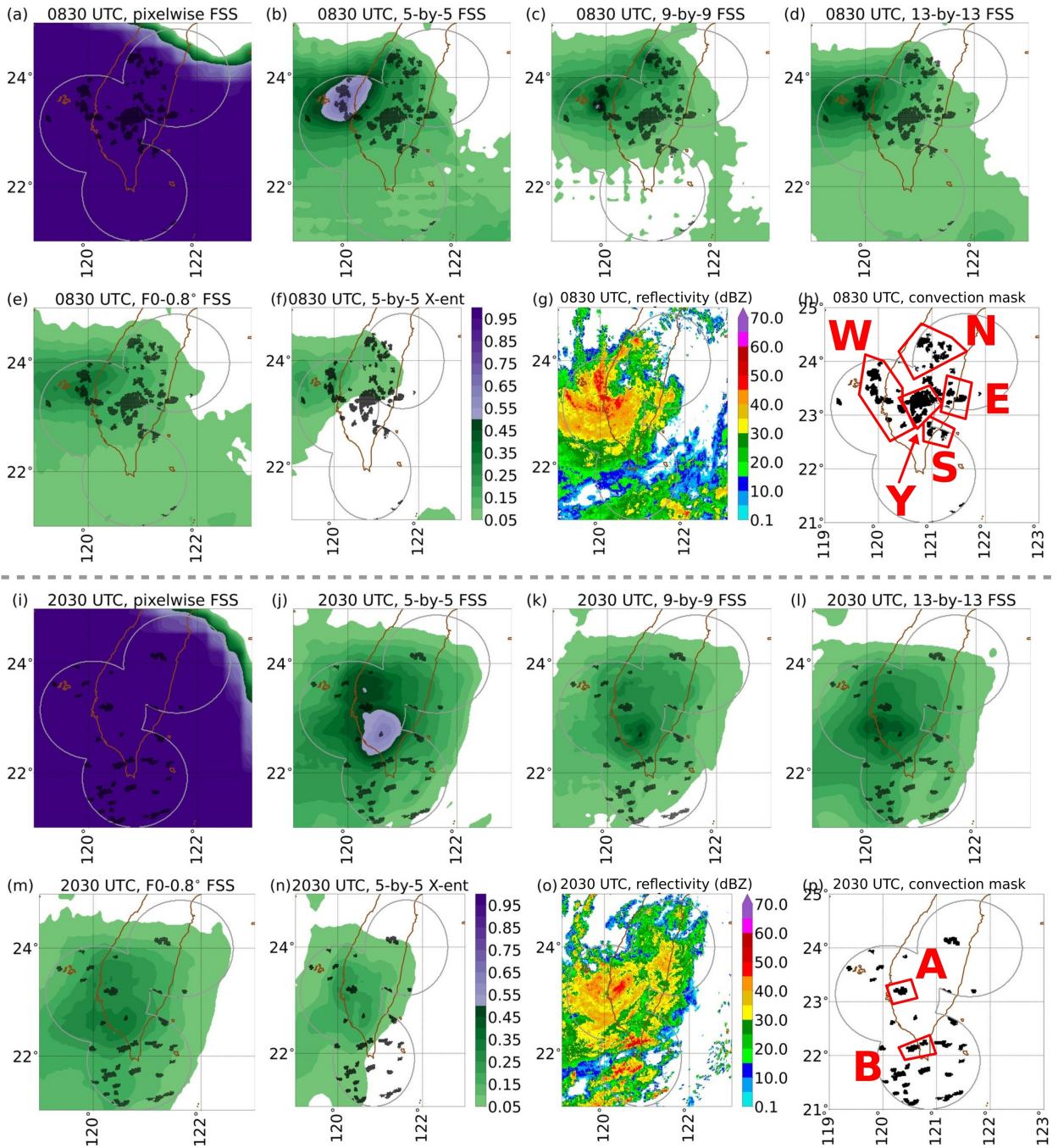

Figure S5: Predictions made by final 2-hour U-nets (each trained with a different loss function) for Tropical Depression Luis. Formatting (black dots and panel titles) is explained in the captions of Figures 10-11 in the main body. [a-f] Forecast convection probabilities valid at 0830 UTC 23 Aug 2018. [g-h] Composite reflectivity and convection mask valid at 0830 UTC 23 Aug 2018. [i-n] Forecast convection probabilities valid at 2030 UTC 23 Aug 2018. [o-p] Composite reflectivity and convection mask valid at 2030 UTC 23 Aug 2018.

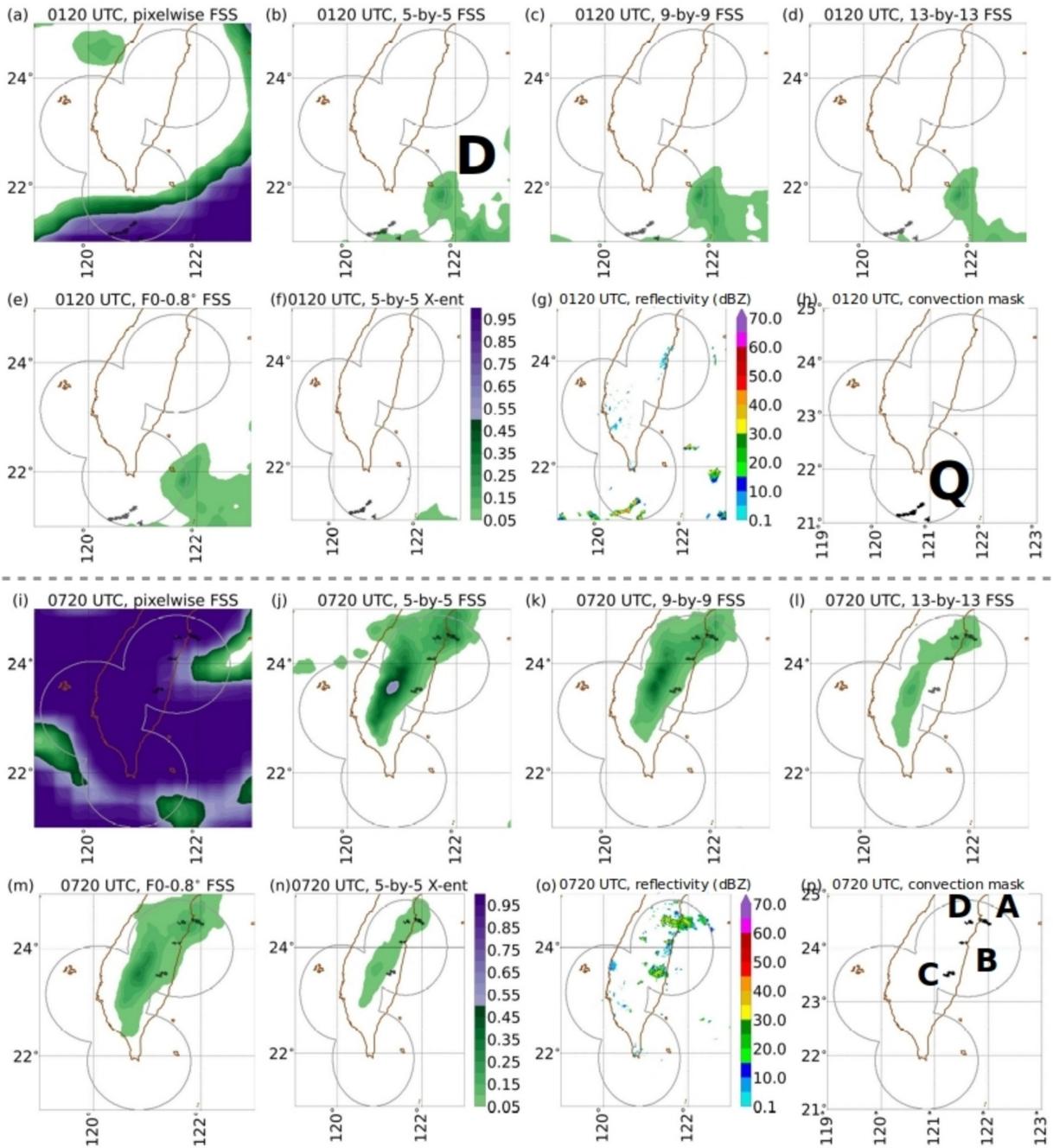

Figure S6: Predictions made by final 2-hour U-nets (each trained with a different loss function) for summer case. Formatting (black dots and panel titles) is explained in the captions of Figures 10-11 in the main body. [a-f] Forecast convection probabilities valid at 0120 UTC 3 Jun 2018. [g-h] Composite reflectivity and convection mask valid at 0120 UTC 3 Jun 2018. [i-n] Forecast convection probabilities valid at 0720 UTC 3 Jun 2018. [o-p] Composite reflectivity and convection mask valid at 0720 UTC 3 Jun 2018.

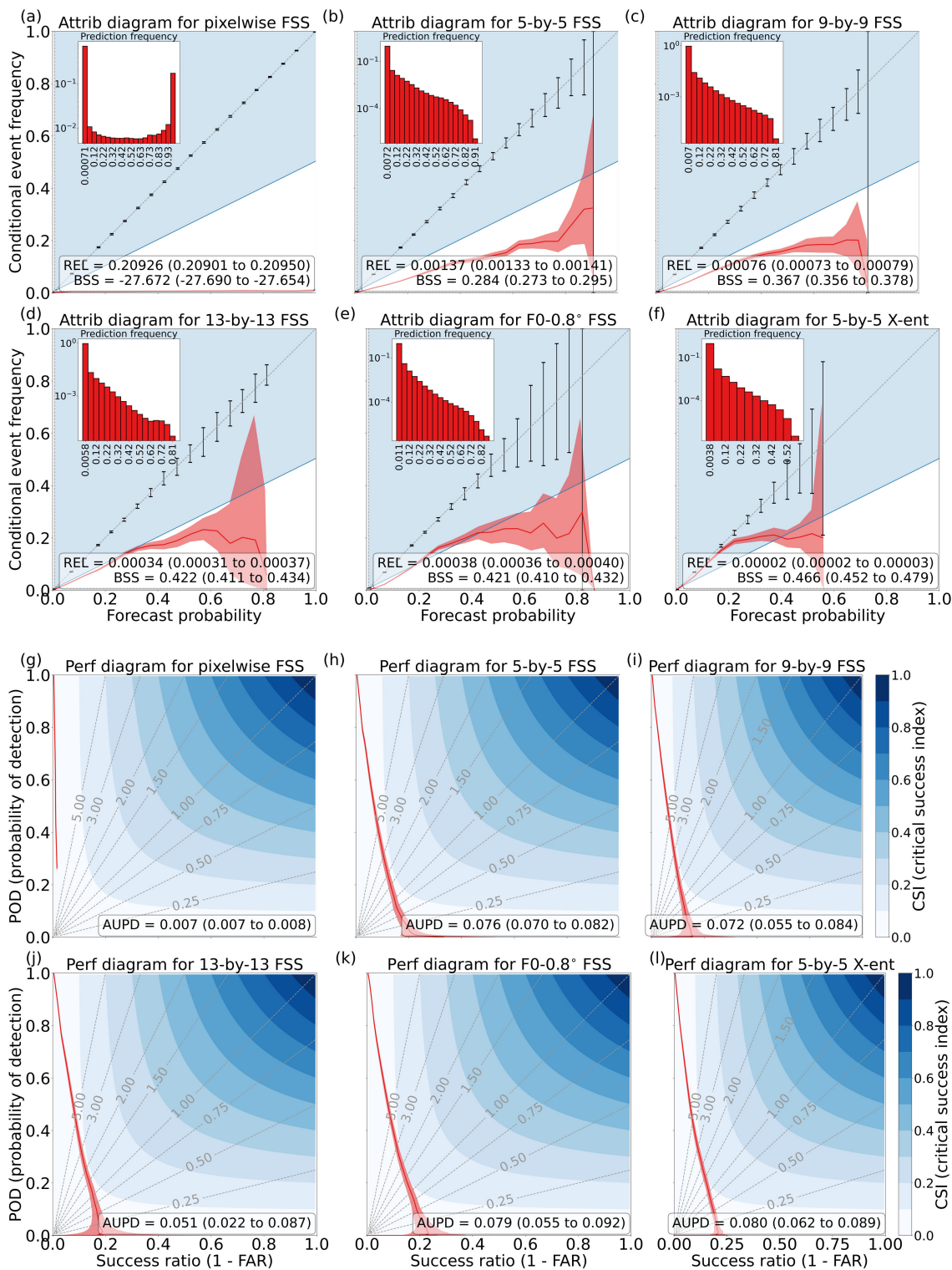

Figure S7: Attributes diagram and performance diagram for each final model. Formatting is explained in the caption of Figure 15 in the main body. [a] Attributes diagram for model trained with pixelwise FSS; [b] 5-by-5 FSS; [c] 9-by-9 FSS; [d] 13-by-13 FSS; [e] F0-0.8° FSS; [f] 5-by-5 cross-entropy. [g-l] Same as a-f but for performance diagrams.

21